%% file: main.tex
\documentclass[10pt,twocolumn,letterpaper]{article}

\usepackage{cvpr}
\usepackage{times}
\usepackage{epsfig}
\usepackage{graphicx}
\usepackage{amsmath}
\usepackage{amssymb}
\usepackage{dsfont}
\usepackage{epstopdf}
\usepackage[abs]{overpic}

\newcommand{\bb}[1]{\bm{\mathrm{#1}}}
\usepackage{bm}

\cvprfinalcopy 

\def\cvprPaperID{95} 

\ifcvprfinal\fi
\begin{document}

\title{Deformable Shape Completion with Graph Convolutional Autoencoders}

\author{Or Litany$^{1,2}$,
        Alex Bronstein$^{1,3}$,
        Michael Bronstein$^{4}$,
        Ameesh Makadia$^{2}$
		\\
		$^1$Tel Aviv University \mbox{   }
		$^2$Google Research \mbox{   }
        $^3$Technion \mbox{   }
        $^4$USI Lugano\mbox{   }
		}

\maketitle

\begin{abstract}
   The availability of affordable and portable depth sensors has made scanning objects and people simpler than ever. However, dealing with occlusions and missing parts is still a significant challenge. The problem of reconstructing a (possibly non-rigidly moving) 3D object from a single or multiple partial scans has received increasing attention in recent years. In this work, we propose a novel learning-based method for the completion of partial shapes. Unlike the majority of existing approaches, our method focuses on objects that can undergo non-rigid deformations. The core of our method is a variational autoencoder with graph convolutional operations that learns a latent space for complete realistic shapes. At inference, we optimize to find the representation in this latent space that best fits the generated shape to the known partial input. The completed shape exhibits a realistic appearance on the unknown part. We show promising results towards the completion of synthetic and real scans of human body and face meshes exhibiting different styles of articulation and partiality.
\end{abstract}


\begin{figure*}
\centering
\includegraphics[width=1\textwidth]{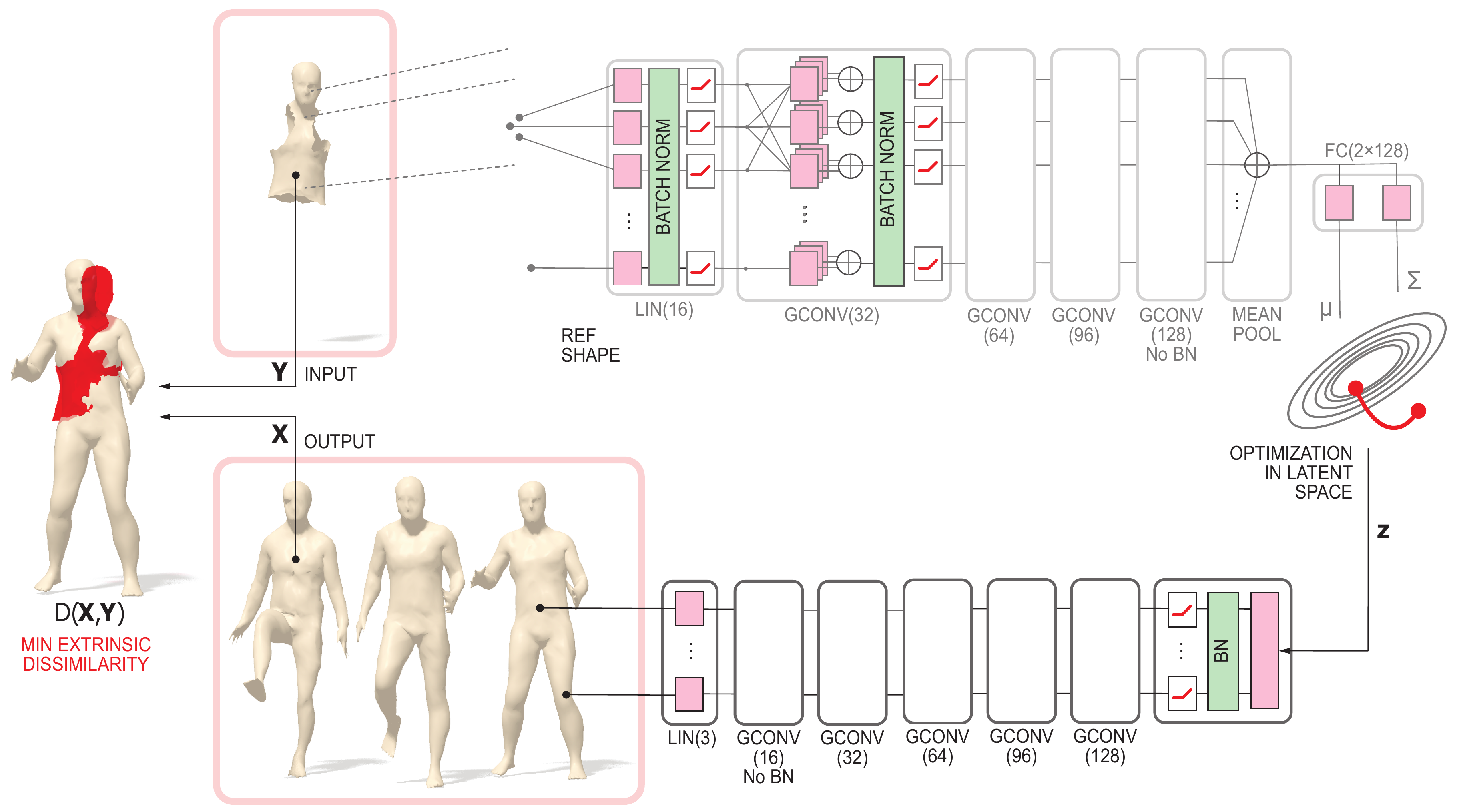}
\caption{\textbf{Schematic description of our approach.} A variational autoencoder is first trained on full shapes with vertex-wise correspondence to create a reference shape and a latent space parameterizing the embedding of its vertices in $\mathbb{R}^3$. At inference, only the decoder (bottom part) is used. A partially missing shape is given as the input together with the correspondence with the reference shape. Starting at a random initialization, optimization is performed in the latent space to minimize the extrinsic dissimilarity between the input shape and the generated output shape.  } 
\label{fig:teaser}
\end{figure*}

\input{./intro.tex}
\input{./related_work.tex}
\input{./method.tex}

\input{./experiments.tex}
%
%
\input{./results.tex}
\input{./conclusions.tex}

\section*{Acknowledgement}
The authors wish to thank Emanuele Rodol\`{a}, Federico Monti, Vikas Sindhwani, and Leonidas Guibas for useful discussions. Much appreciated is the DFAUST scan data provided by Federica Bogo and Senya Polikovsky.

\clearpage

{\small
\bibliographystyle{ieee}
\bibliography{egbib}
}


\end{document}

%% file: intro.tex
\section{Introduction}
The problem of reconstructing 3D shapes from partial observations is central to a broad spectrum of applications, ranging from virtual and augmented reality to robotics and autonomous navigation. Of particular interest is the setting where objects may undergo articulations or more generally non-rigid deformations. While several methods based on (volumetric) convolutional neural networks have been proposed for completing man-made rigid objects (see~\cite{dai2016shape,song2016semantic,varley17iros,wu20153d,sharma16eccvw}), they struggle at handling deformable shapes. However, this is not a limitation specific to volumetric approaches. The same difficulties with deformable shapes, irrespective of the completion task, are present for other 3D shape representations utilized in deep learning frameworks, such as view-based~\cite{su2015multi,wei2016dense} and point clouds~\cite{qi2016pointnet,qi2017pointnet++}.

The main reason for this is that for methods based on Euclidean convolutional operations (e.g. volumetric or view-based deep neural networks), an assumption of self-similarity under rigid transformations (in most cases, axis-aligned) is implied. For example a chair seat will always be parallel to the floor. Non-rigid deformations violate this assumption, effectively making each pose a novel object. Thus, tackling such data with a standard CNN requires many network parameters and a prohibitively large amount of training. Although model-based methods such as \cite{anguelov2005scape} have shown good performance, they are restricted to a specific class of shape with manually constructed models. 

To explicitly enable robustness towards non-rigid deformations, the approach advocated in this paper adopts recent advances for in CNNs on graphs which directly exploit the 3D mesh structure. This allows the learning of a powerful non-rigid shape representation from data without an explicit model.

Another shortcoming of deep learning shape completion techniques stems from their end-to-end design. A network trained to perform completion would be biased towards the type of missing data introduced at training, and may not generalize well to previously unseen types of missing information. To allow generalization to any style of partiality we choose to separate the task of completion from the training procedure altogether. As a result, we also avoid a significant amount of preprocessing and augmentation that is typically done on the training data.

Finally, when a complete mesh is desired as the output, producing a triangulation from point clouds or volumetric grids is itself a challenging problem and may introduce undesired artifacts (although recent advances such as~\cite{dai2016shape} address this by directly producing implicit surfaces). Conversely, by utilizing a mesh-convolutional network our method will produce complete and plausible surfaces by design.  

\paragraph{Contribution.} The main contribution of this work is a method for deformable shape completion that decouples the task of partial shape completion from the task of learning a generative shape model, for which we introduce a novel graph convolutional autoencoder architecture. Compared to previous works the proposed method has several advantages. First, it can handle any style of partiality without needing to see any partial shapes during training. Second, the method is not limited to a specific class of shapes (e.g. humans) and can be applied to any kind of 3D data. Third, shape completion is an inherently ill-posed problem with potentially multiple valid solutions fitting the data (this is especially true for articulated and deformable shapes), thus making deterministic solutions inadequate. The proposed method reflects the inherent ambiguities of the problem by producing multiple plausible solutions.

%% file: related_work.tex
\section{Related work}
\label{sec:relatedwork}
\paragraph{3D shape completion.} 
The application addressed in this paper is a very active research area in computer vision and graphics, ranging from completion of small holes~\cite{sarkar2017learning} and larger missing regions in individual objects~\cite{pointcloudGAN,rock2015completing,varley17iros,wu20153d,sharma16eccvw}, to entire scenes~\cite{song2016semantic}.
Completion guided by geometric priors has been explored, for example Poisson filling~\cite{kazhdan2013screened} and self-similarity~\cite{korman2015peeking,sarkar2017learning,litany2016cloud}. However, such methods work only for small missing regions, and dealing with bigger occlusions requires stronger priors. A viable alternative is model-based approaches, where a parametric morphable model describing the variability of a certain class of objects can be fit to the observed data~\cite{blanz1999morphable,gerig2017morphable}. 

The setting of non-rigid shape completion differs from its rigid counterpart in that at inference, the input partial shape may admit a deformation unseen in the training data. This distinction becomes crucial as large missing regions force the priors to become more complex (see for example the human model designed in \cite{anguelov2005scape}).

\paragraph{Generative methods for non-rigid shapes.} 
The state-of-the-art in generative modeling has rapidly advanced with the introduction of Variational Autoencoders (VAE~\cite{kingma2014iclr}), Generative Adversarial Networks~\cite{goodfellow2014generative}, and related variations (e.g. VAEGAN~\cite{larsen16icml}). These advances have been adopted by the 3D shape analysis community for dynamic surface generation through VAE~\cite{kostrikov2017surfnet} and image-to-shape generation through VAEGAN~\cite{wu16nips}. In~\cite{tan2017variational}, a VAE for non-rigid shapes is proposed. This work differs from ours in that the core operations of our network are graph-convolutional operations as opposed to fully-connected layers, and our network operates directly on raw 3D vertex positions rather than relying on hand-crafted features.

\paragraph{Geometric deep learning.} 
This paper is closely related to a broad area of active research in geometric deep learning (see~\cite{gdl} for a summary). The success of deep learning (in particular, convolutional architectures \cite{lecun1998gradient}) in computer vision has brought a keen interest in the computer graphics community to replicate this progress for applications dealing with geometric 3D data. One of the key difficulties is that for such data it requires great care to define the basic operations constituting deep neural networks, such as convolution and pooling.

Several works avoid this problem by using a Euclidean representation of 3D shapes, such as rendering a collection of 2D views \cite{su2015multi,wei2016dense}, volumetric representations \cite{wu20153d}, or point cloud \cite{qi2016pointnet,qi2017pointnet++}. One of the main drawbacks of such extrinsic deep learning methods is their difficulty to deal with shape deformations as discussed earlier. Additionally, voxel representations are often memory intensive and suffer from poor resolution \cite{wu20153d}, although recent models have been proposed to address these issues: implicit surface representation~\cite{dai2016shape}, sparse octree networks~\cite{wang17ocnn,riegler17cvpr}, encoder-decoder CNN for patch-level geometry refinement~\cite{han17iccv}, and a long-term recurrent CNN for upsampling coarse shapes~\cite{wang17iccv}. Regarding point cloud representations, the PointNet model~\cite{qi2016pointnet} applies identical operations to the coordinates of each point and aggregates this local information without allowing for interaction between different points which makes it difficult to capture local surface properties. PointNet++~\cite{qi2017pointnet++} addresses this by proposing a spatially hierarchical model. Additionally, for PointNet to be invariant to rigid transformations the input point clouds are aligned to a canonical space. This is achieved by a small network that predicts the appropriate affine transformation, but in general such an alignment would be difficult for articulated and deformable shapes.

An alternative strategy is to redefine the basic ingredients of deep neural networks in a geometrically meaningful or intrinsic manner. The first intrinsic CNN-type architectures for 3D shapes were based on local charting techniques generalizing the notion of ``patches'' to non-Euclidean and irregularly-sampled domains \cite{masci15,boscaini2016learning,monet}. The key advantage of this approach is that the generalized convolution operations are defined intrinsically on the manifold, and thus automatically invariant to its isometric deformations. As a result, intrinsic CNNs are capable of achieving correspondence results with significantly less parameters and a very small training set. Related independent efforts developed CNN-type architectures for general graphs \cite{bruna2013spectral,henaff2015deep,defferrard2016convolutional,kipf2016semi,monet,levie2017cayleynets}. 

Recently,~\cite{dynFilt} suggested a dynamic filter in which the assignment of each filter to each member of the k-ring in a graph neighborhood is determined by its feature values. Importantly, this method demonstrated state-of-the-art performance working directly on the embedding features. Thus, in our work we build upon~\cite{dynFilt} as a basic building block for convolution operations. 

\paragraph{Partial shape correspondences.}
Dense non-rigid shape correspondence \cite{kim11,chen15,fmnet,rodola14,bronstein2006generalized} is a fundamental challenge as it is an enabler for many high level tasks like pose or texture transfer across surfaces. We refer the interested reader to \cite{van2011survey,biasotti2015recent} for a detailed review of the literature. The proposed method in this work builds upon correspondence between a partial input and a canonical shape of the same class, and related to this are several methods that explore partial shape correspondence and matching~\cite{rodola16-partial,monet,litany17fully}. The approaches demonstrating state-of-the-art performance on partial human shapes (e.g.~\cite{monet}) treat correspondence as a vertex classification task. Recently~\cite{wei2016dense} has shown impressive results for correspondence across different human subjects in varied pose and clothing. 

\paragraph{Inpainting.}
The 3D shape completion task is closely related to the analogous structured prediction task of image inpainting~\cite{pathakCVPR16context,Yang_2017_CVPR}. However, our proposed optimization scheme is more reminiscent of style transfer~\cite{gatysStyleTransfer} techniques. In our setting we optimize only for the best complete shape with no constraints on the internal feature representation.

%% file: method.tex
\section{Method}
\label{sec:method}
We propose a shape completion method that detaches the process of learning to generate 3D shapes from the task of partial shape completion. Our method requires a generative model for complete 3D shapes which we construct by training a graph-convolutional variational autoencoder (VAE~\cite{kingma2014iclr}). Partial shapes can be completed by identifying the shape in the output space of the VAE's generator which best aligns with the partial input. We propose an optimization in the latent space that iteratively deforms (non-rigidly) a randomly generated shape to align with a partial input. In what follows, we describe in more detail both ingredients of our process, the VAE generator and the partial shape completion scheme. A schematic rendition of the method is depicted in Figure \ref{fig:teaser}.

\paragraph{3D shape generator.}
We fix the number of vertices $N$ and the topology of a reference shape and refer to the three-dimensional vertex embedding $\bb{X}  \in \mathbb{R}^{3 \times N}$ as to a shape.
The VAE consists of two networks: the \emph{encoder} that encodes 3D shape inputs $\bb{X}$ to a latent representation vector $\bb{z}=\mathrm{enc}(\bb{x})$, and the \emph{decoder} that decodes the latent vectors into 3D shapes $\bb{X}'=\mathrm{dec}(\bb{z})$. The variational distribution $q(\bb{z}|\bb{X})$ is associated with a prior distribution over the latent variables, and the usual choice which we follow here is a centered multivariate Gaussian with unit variance $\mathcal{N}(\bb{0},\bb{I})$. Our VAE loss combines the shape reconstruction loss $L_r = ||\mathrm{dec}\circ \mathrm{enc} (\bb{X})-\bb{X}||_2$ encouraging the encoder-decoder pair to be a nearly identity transformation, and a regularization prior loss measured by the Kullback-Leibler divergence, $L_p=D_\mathrm{KL}(q(\bb{z}|\bb{X})||p(\bb{z}))$. The total VAE loss is computed as $L=L_r + \lambda L_p$, where $\lambda \geq 0$ controls the similarity of the variational distribution to the prior.

The choice to measure shape reconstruction loss with pointwise distances is not the only option. For example, the VAE can be combined with a Generative Adversarial Network (VAE-GAN) as in~\cite{larsen16icml,wu16nips}, thus introducing an additional discriminator loss on the reconstructed shape. We do not consider a discriminator in the scope of this work to avoid additional model complexity but leave it as future work to investigate different loss functions that can be imposed on reconstructed shapes.

The internal details of the VAE encoder $\mathrm{enc}(\bb{X})$ and decoder $\mathrm{dec}(\bb{z})$ are largely influenced by the choice of the 3D shape representation. As discussed in Section~\ref{sec:relatedwork}, many representations have been explored ranging from voxels to raw point clouds. Our desire to focus on shape completion for deformable object classes leads us to consider intrinsic mesh and surface models that have shown promising results for deformable shape correspondence among other applications (e.g.~\cite{masci15,monet}). Multiple approaches have been proposed to perform convolution on spatial meshes. The primary factor which distinguishes spatial graph convolutional operations is how correspondence is determined between convolutional filters and the local graph neighborhoods. Rather than relying on properties of the underlying geometry to map filters to surface patches, we adopt data-adaptive models which learn the mapping from the neighborhood patch to filters weights. Specifically, our VAE is primarily composed of the dynamic filtering convolutional layers proposed in FeaStNet~\cite{dynFilt}. The input to the layer is a feature vector field on the mesh vertices, attaching to a vertex $i$ a vector $\bb{x}_i$. The output is also a vector field $\bb{y}_i$, possibly of a different dimension, computed as  
\begin{align}\label{eq:filt}
\bb{y}_i = \bb{b} + \sum_{m=1}^{M} \frac{1}{|\mathcal{N}_i|} \sum_{j\in \mathcal{N}_i} q_m(\bb{x}_i, \bb{x}_j) \bb{W}_m \bb{x}_j,
\end{align}
where $\mathcal{N}_i$ denotes a patch around the vertex $i$, and 
$q_m(\bb{x}_i, \bb{x}_j) \propto \exp (\bb{u}_m^\mathrm{T} (\bb{x}_i - \bb{x}_j) + c_m)$
are positive edge weights in the patch normalized to sum to one over $m$.
The trainable weights of the layer are $\bb{W}_m$, $\bb{u}_m$, $c_m$ and $\bb{b}$, while the number of weight matrices $M$ is a fixed design parameter. Note that the mapping from neighborhood patch to weights is translation invariant in the input feature space, as $q$ operates only on the differences $\bb{x}_i-\bb{x}_j$. Refer to Figure \ref{fig:teaser} and \cite{dynFilt} for further details.

\paragraph{Partial shape completion.}
Once the encoder-decoder pair has been trained, the encoder is essentially tossed away, while the decoder acts as a complete shape generator, associating to each input latent vector $\bb{z}$ an $\mathbb{R}^3$ embedding of the reference shape, $\bb{X} = \mathrm{dec}(\bb{z})$. Importantly, this acts as a strong shape prior, generating plausible looking shapes (see Figure~\ref{fig:random_sample}).

At inference, a partial shape $\bb{Y}$ is given. We first use an off-the-shelf method (MoNet) \cite{monet} to compute a dense  partial intrinsic correspondence between $\bb{Y}$ and the reference shape. Representing this correspondence as a partial permutation matrix $\bb{\Pi}$ and applying it to any shape $\bb{X}$ generated by the decoder produces a subset of points in $\mathbb{R}^3$, $\bb{X} \bb{\Pi} )$, ordered compatibly with their counterparts in $\bb{Y}$. 
We therefore define an extrinsic dissimilarity between the input shape and the generated full shape as
$\mathrm{D}(\bb{X},\bb{Y}) =  \| \bb{X} - \bb{Y} \|$, possibly weighed by the confidence of the correspondence at each point. 

Inference consists essentially of finding a latent vector $\bb{z}^\ast$ minimizing the dissimilarity between the input and the output shape,
\begin{eqnarray}
\min_{\bb{z}, \bb{T} \in \mathrm{SE}(3)} \mathrm{D}(\mathrm{dec}(\bb{z}) \bb{\Pi}, \bb{T} \bb{Y} ),
\label{eq:optimization}
\end{eqnarray}
where $\bb{T}$ denotes a rigid transformation. Alternating steps are performed over $\bb{z}$ (non-rigid deformation) and $\bb{T}$ (rigid registration). When the $\ell_2$ norm is used to define the shape dissimilarity, the rigid registration step has a closed-form solution via the singular value decomposition of the covariance matrix of $\bb{Y}$ and $\bb{X}\bb{\Pi}$, while the non-rigid deformation step is performed using stochastic gradient descent.

Shape completion is an inherently ill-posed problem that can have multiple plausible solutions. In cases where there exists more than one solution consistent with the data, sampling a result from our proposed generative model allows us to explore this space. The results in Section \ref{subsec:completion_variability} illustrate the variability in completed shapes when repeating the optimization procedure (\ref{eq:optimization}) with random initializations.

%% file: experiments.tex
\section{Experiments}
\paragraph{Dataset.} 
The majority of our experiments are performed on human shapes. The VAE is trained on registered 4D scans from the DFAUST dataset~\cite{dfaust:CVPR:2017} comprising $10$ human subjects performing $14$ different activities. Scans are captured at a high frame rate and registered to a canonical topology. Due to the high frame rate, we subsample the data temporally by a factor of $4$. We consistently subsample each mesh by the factor of $2$ down to $N=3446$ vertices. Refer to the supplemental info for details on the data processing. The training set is created by holding out all scans for two human subjects and two activities leaving approximately $7000$ training shapes. Details for additional experiments with face meshes is provided in Section~\ref{sec:facecompletion}.

\paragraph{Network parameters.} 
The structure of our graph-convolutional VAE is illustrated in Figure~\ref{fig:teaser}. We evaluated a number of model parameters on a subset of the training set to inform our final design choices. We use $M=8$ and latent dimensionality of $128$ for all our DFAUST experiments.  
A more important and delicate decision is the selection of the parameter $\lambda$ controlling the emphasis on pushing the variational latent distribution towards the Gaussian prior. Our experiments show, as expected, that a higher weight for the Gaussian prior causes randomly sampled latent vectors to generate realistic shapes more likely, while a lower $\lambda$ improves reconstruction accuracy over a wider variety of shapes. In the context of our problem, it is more important for the latent space to represent and for the decoder to be able to generate a wide variety of shapes accurately. Sampling from the latent space is less important since the final latent vectors are obtained by means of solving the optimization problem (\ref{eq:optimization}). Consequently, we selected $\lambda = 10^{-8}$ at training (see supplemental material for empirical analysis motivating these choices).

\paragraph{Implementation details.}
We train the model directly on the $3 \times N$ input meshes from the DFAUST dataset as described above; the sparse adjacency matrices (we use a vertex 2-ring as the neighborhood size) are passed as side information to define the graph convolutional layers. Data are augmented by adding normally distributed noise to the vertex positions as well as a global planar translations and scalings. We use the ADAM~\cite{kingma2014adam} optimizer with the learning rate set to $10^{-4}$, momentum to $0.9$, batch size to $2$, Xavier initialization~\cite{xavierInit} for all weights, and train for $3 \times 10^5$ iterations. For shape completion optimization we use an SGD optimizer with a $0.1$ learning rate. All additional data for training and evaluation will be provided on the authors' websites.

%% file: results.tex
\subsection{Representation quality}
\begin{figure}
\centering
\includegraphics[height=0.15\textwidth]{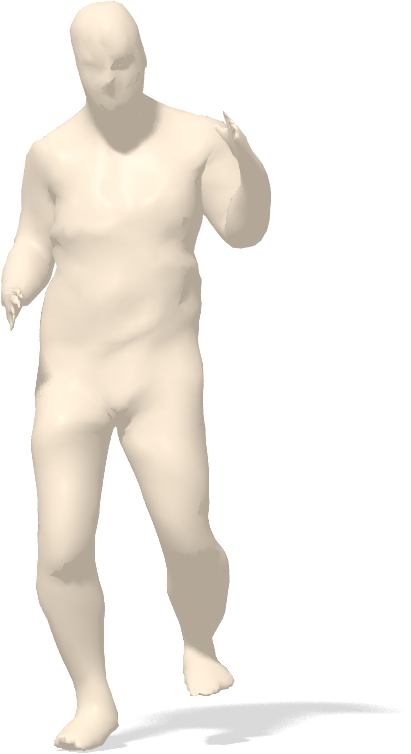}
\includegraphics[height=0.15\textwidth]{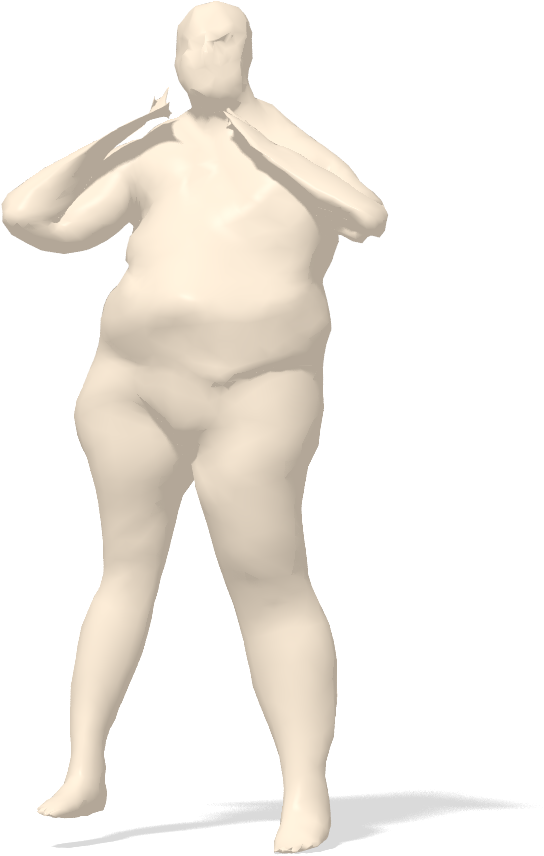}
\includegraphics[height=0.15\textwidth]{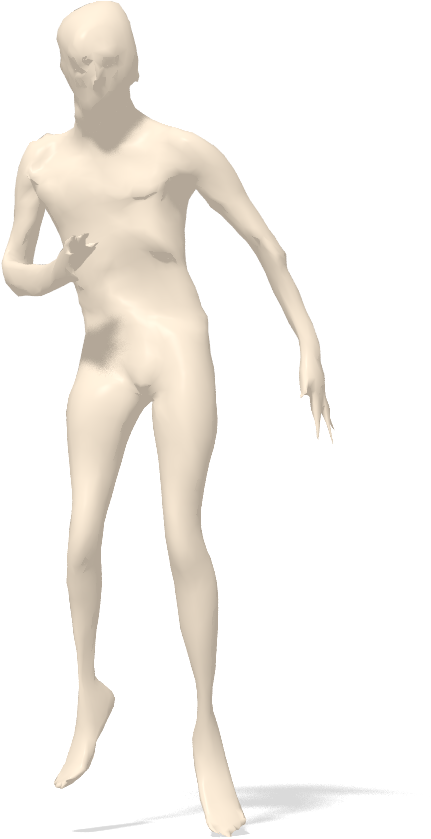}
\includegraphics[height=0.15\textwidth]{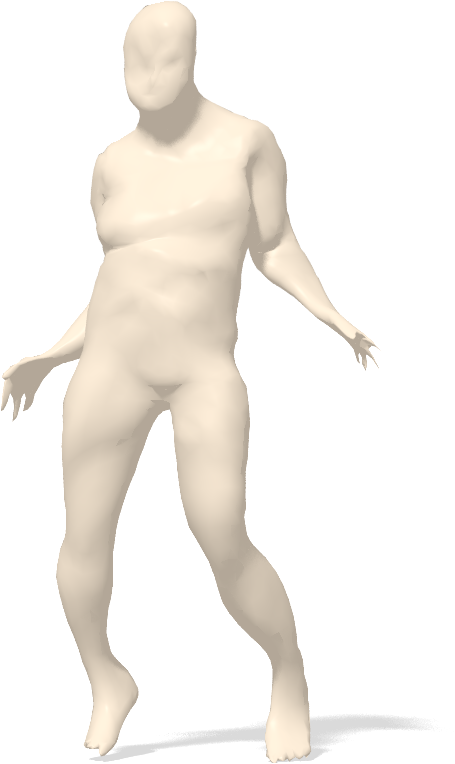}
\includegraphics[height=0.15\textwidth]{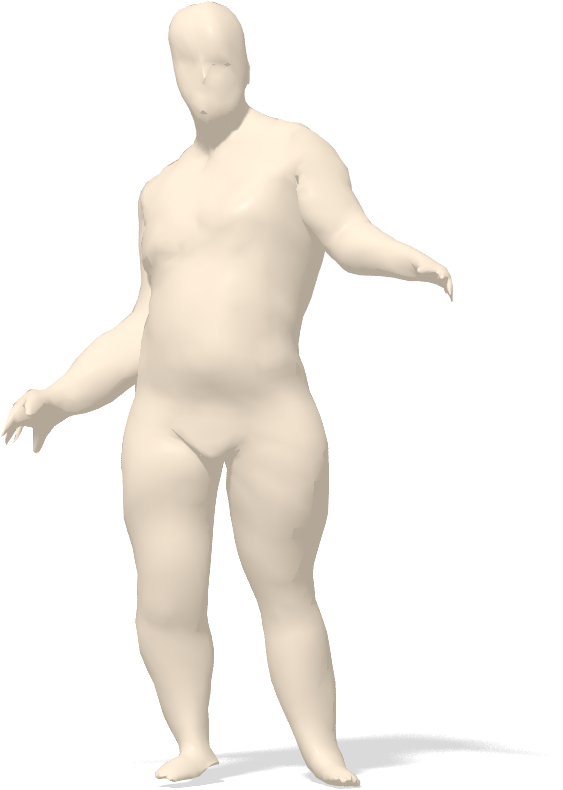}
\includegraphics[height=0.15\textwidth]{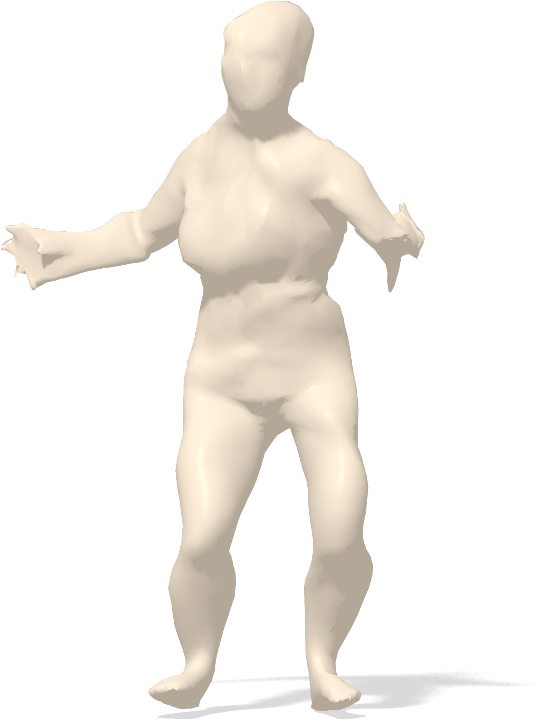}
\includegraphics[height=0.15\textwidth]{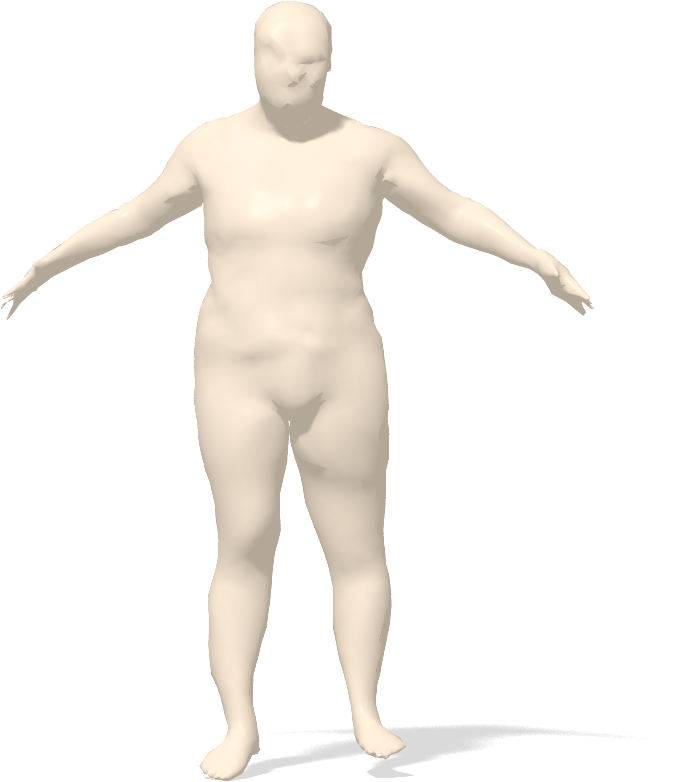}
\includegraphics[height=0.15\textwidth]{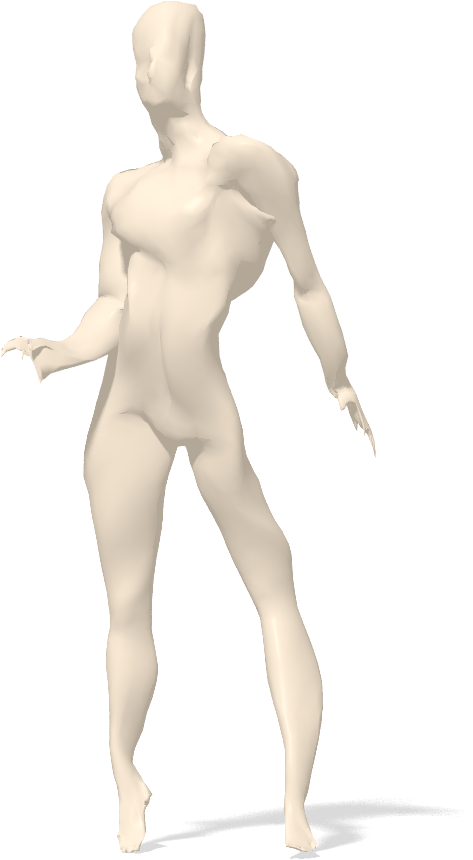}
\includegraphics[height=0.15\textwidth]{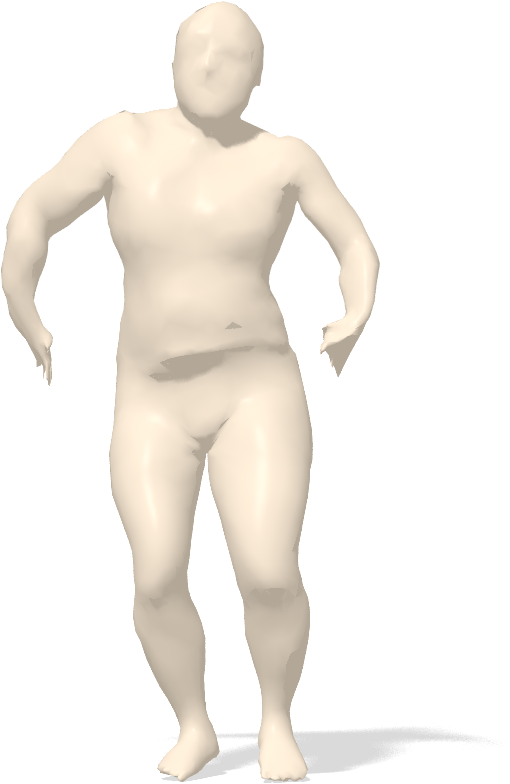}

\caption{\textbf{Random human shapes generated by the VAE.} We have explicitly relaxed the Gaussian prior on the latent variables during training. The tradeoff is that samples coming from the prior may generate slightly unrealistic shapes.}
\label{fig:random_sample}
\end{figure}

To understand the generative capabilities of the VAE we show several examples related to shape generation as well as explore the structure of the learned latent space.
Figure~\ref{fig:random_sample} depicts shapes generated by the decoder fed with latent variables randomly sampled from $\mathcal{N}(\bb{0},\bb{I})$. As we have explicitly relaxed the Gaussian prior on the latent variables (small $\lambda$) during training. As discussed earlier, the tradeoff is that samples coming from the prior may generate slightly unrealistic shapes.

Figure \ref{fig:interpolation} depicts generated shapes as the result of linear interpolation in the latent space. The source and target shapes are first passed through the encoder to obtain latent representations; applying the decoder to convex combinations of these latent vectors produces a highly non-linear interpolation in $\mathbb{R}^3$. The top two rows of Figure~\ref{fig:interpolation} show interpolation for networks trained with $\lambda=10^{-6}$ and $\lambda=10^{-8}$, respectively. The bottom row of Figure~\ref{fig:interpolation} highlights the interesting structure of the learned latent space through arithmetic. Applying the difference of a subject with left knee raised and lowered to the same subject with right knee raised results in a lowering of the right knee. The network learned this symmetry without any explicit modeling.

\definecolor{mygray}{rgb}{0.6,0.6,0.6}
\begin{figure}
\centering
\resizebox{1\columnwidth}{!}{
\addtolength{\tabcolsep}{-4pt}
\begin{tabular}{cccccc}
\includegraphics[height=0.12\textwidth]{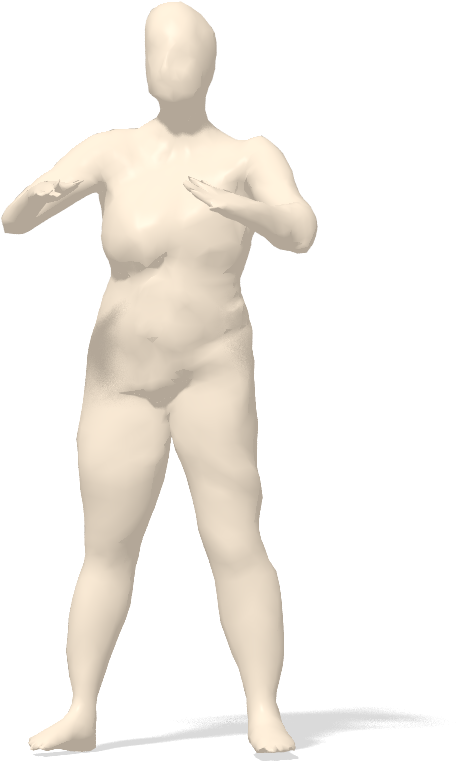} &
\includegraphics[height=0.12\textwidth]{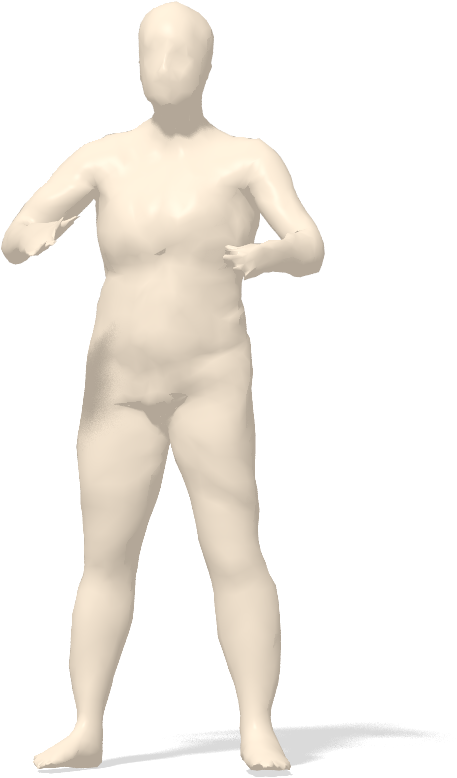} &
\includegraphics[height=0.12\textwidth]{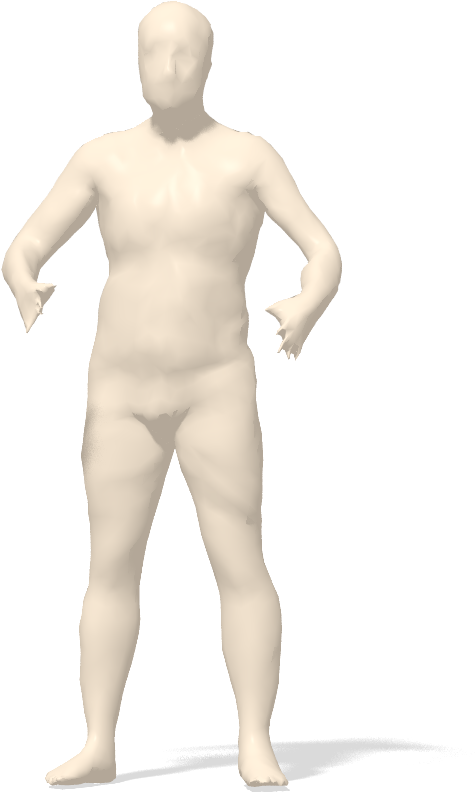} &
\includegraphics[height=0.12\textwidth]{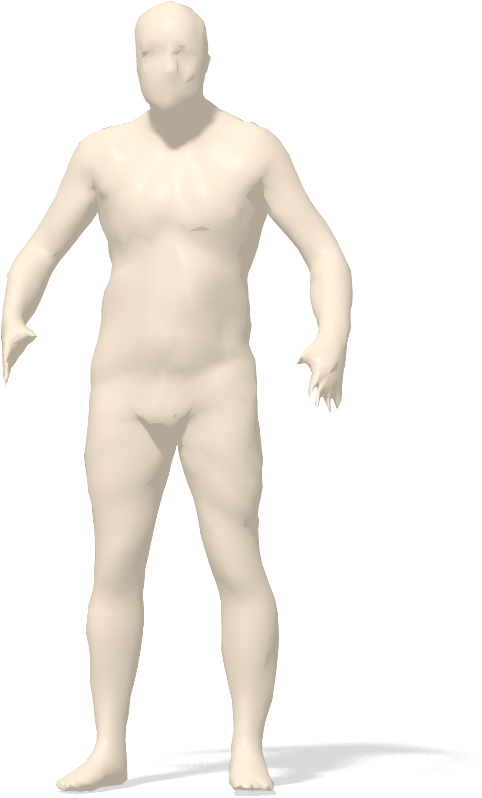} &
\includegraphics[height=0.12\textwidth]{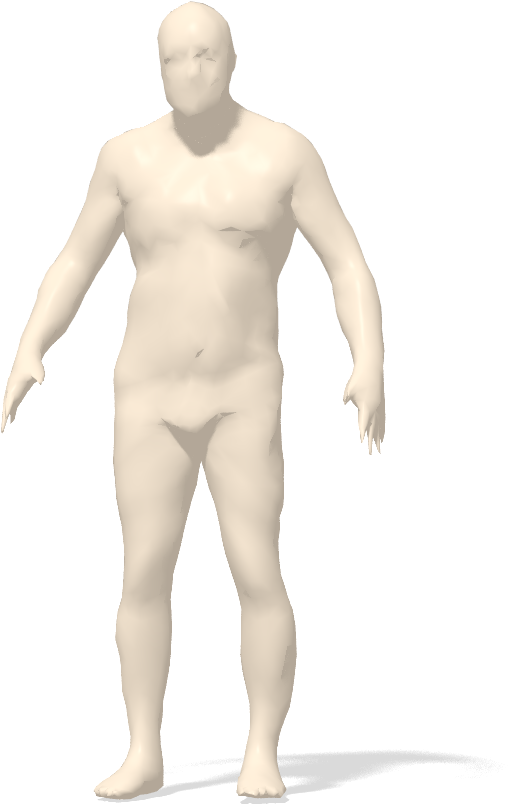} &
\includegraphics[height=0.12\textwidth]{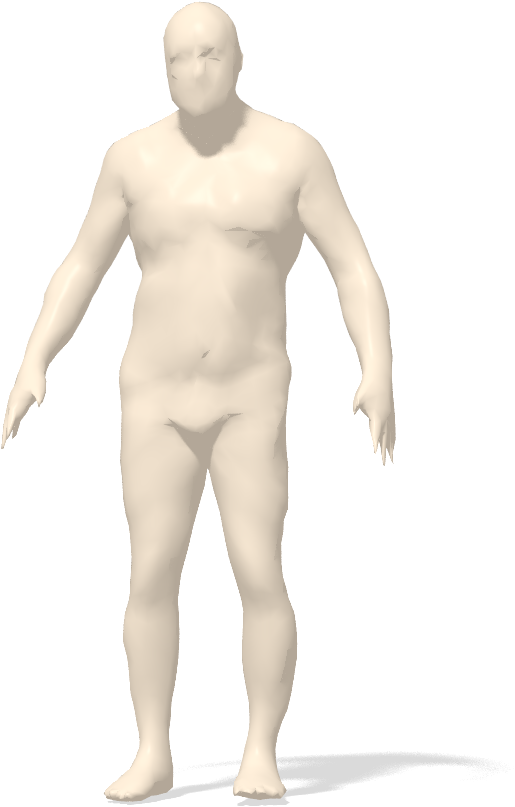} \\
\includegraphics[height=0.12\textwidth]{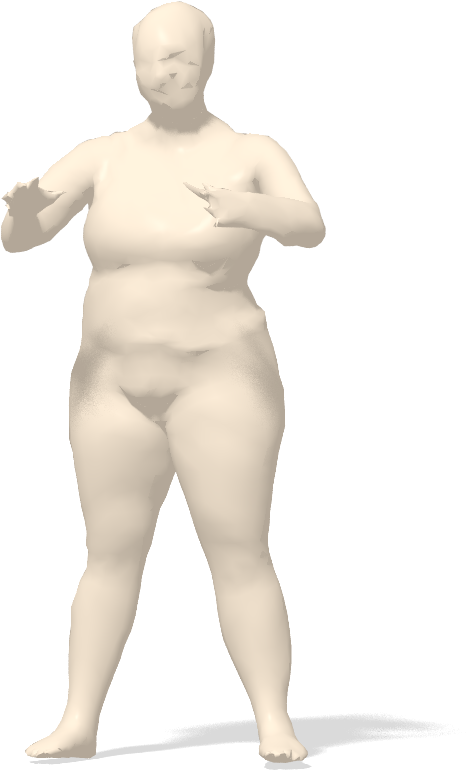} &
\includegraphics[height=0.12\textwidth]{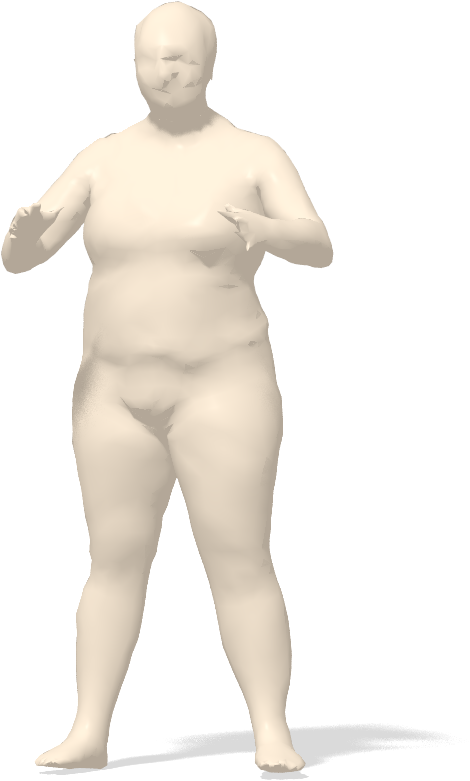} &
\includegraphics[height=0.12\textwidth]{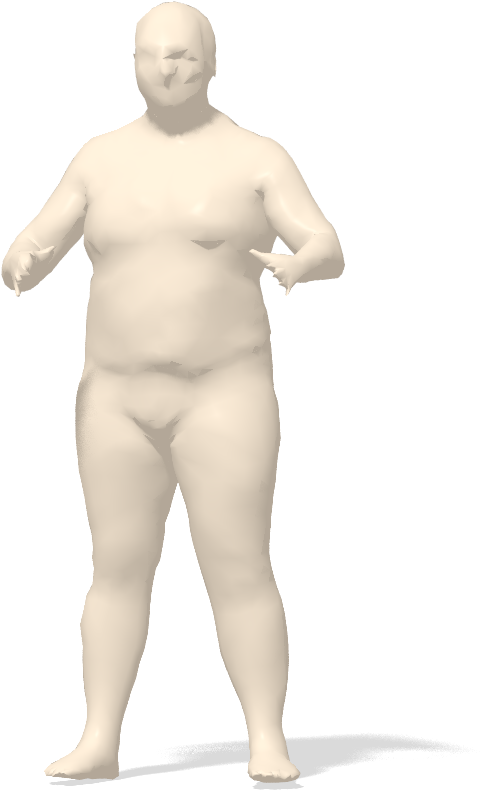} &
\includegraphics[height=0.12\textwidth]{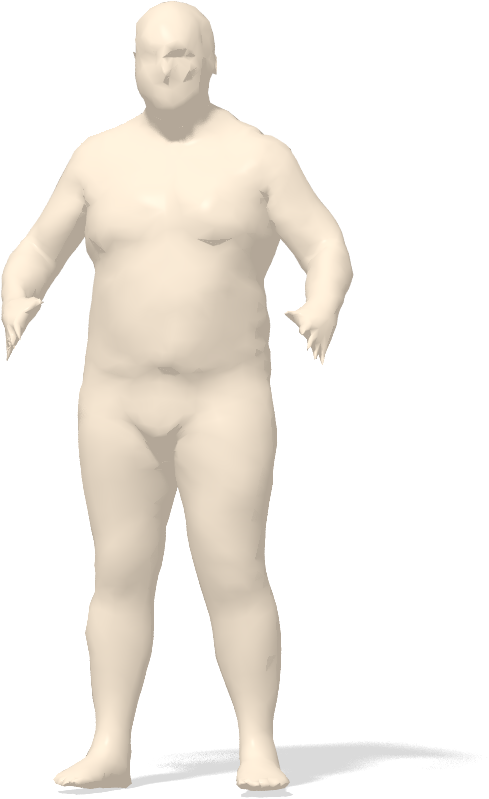} &
\includegraphics[height=0.12\textwidth]{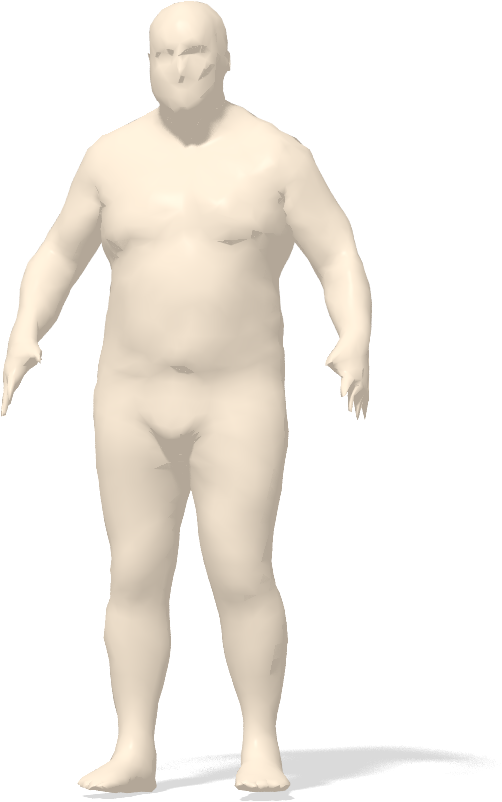} &
\includegraphics[height=0.12\textwidth]{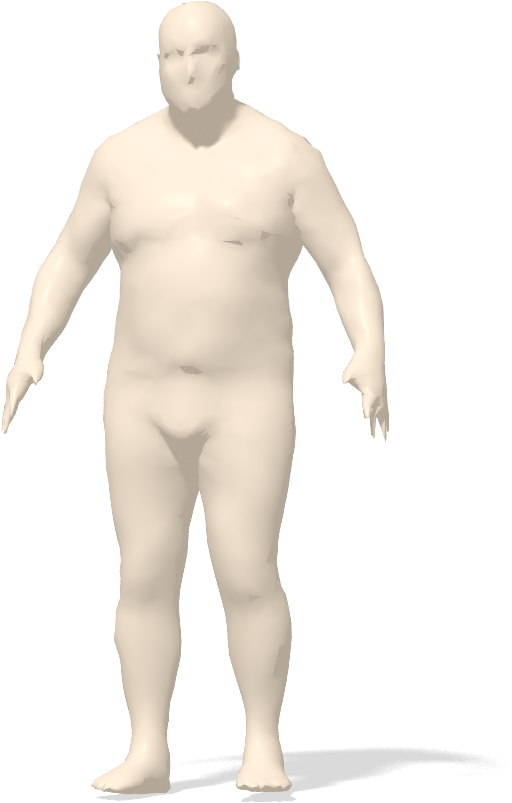} \\ 
\vspace{10pt} \\
\begin{overpic}
[trim=0cm 0cm 0cm 0cm,clip,height=0.12\textwidth]{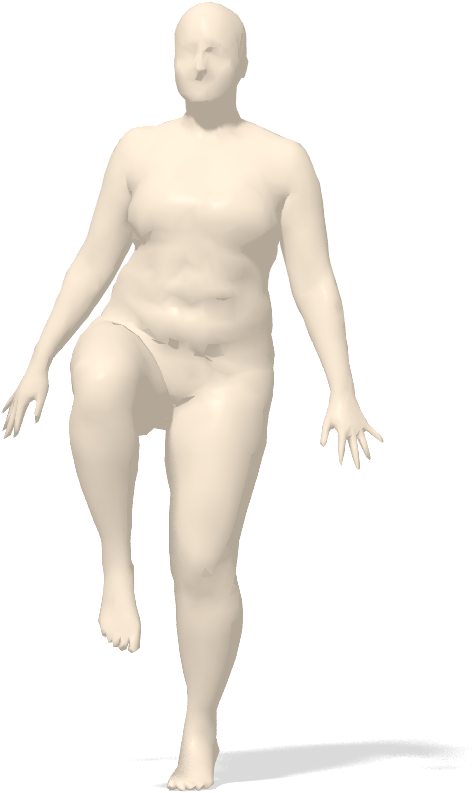}
\put(12,65){\small Z}
\put(23,65){$+$}
\put(33,65){\small $\alpha$}
\put(41,65){$($}
\end{overpic} &
\begin{overpic}
[trim=0cm 0cm 0cm 0cm,clip,height=0.12\textwidth]{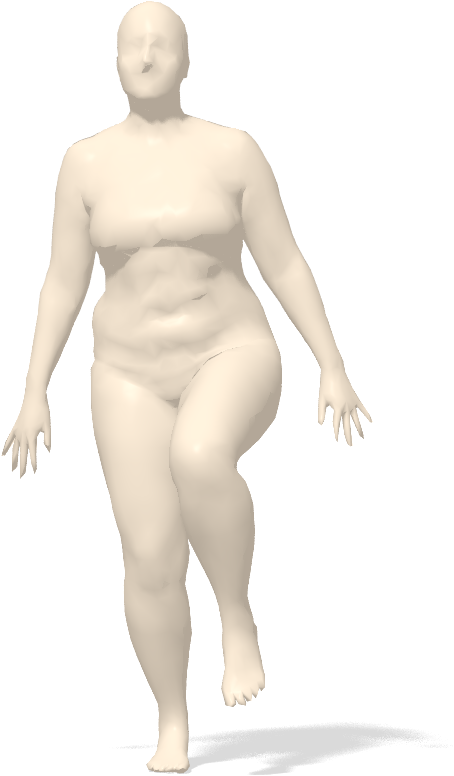}
\put(8,65){\small X}
\put(28,65){$ - $}
\end{overpic} &
\begin{overpic}
[trim=0cm 0cm 0cm 0cm,clip,height=0.12\textwidth]{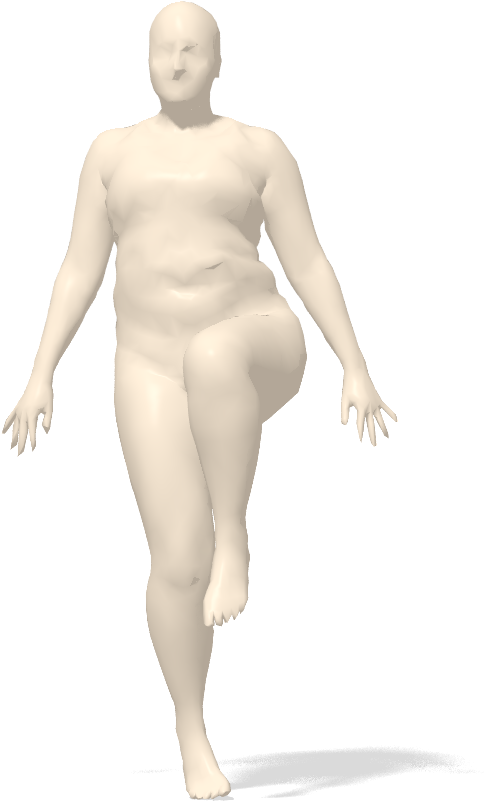}
\put(10,65){\small Y}
\put(20,65){$)$}
\put(35,70){ {\color{mygray}\line(0,-1){70}} }
\end{overpic} &
\begin{overpic}
[trim=0cm 0cm 0cm 0cm,clip,height=0.12\textwidth]{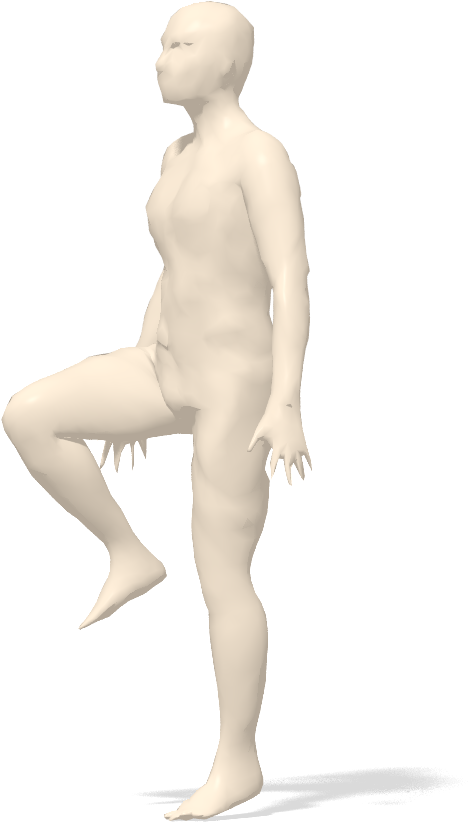}
\put(0,65){\small$\alpha = 0.5$}
\end{overpic} &
\begin{overpic}
[trim=0cm 0cm 0cm 0cm,clip,height=0.12\textwidth]{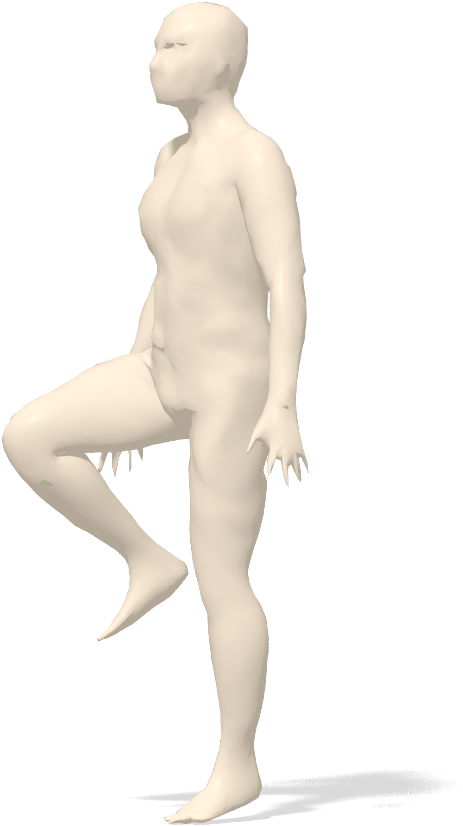}
\put(0,65){\small$\alpha = 0.7$}
\end{overpic} &
\begin{overpic}
[trim=0cm 0cm 0cm 0cm,clip,height=0.12\textwidth]{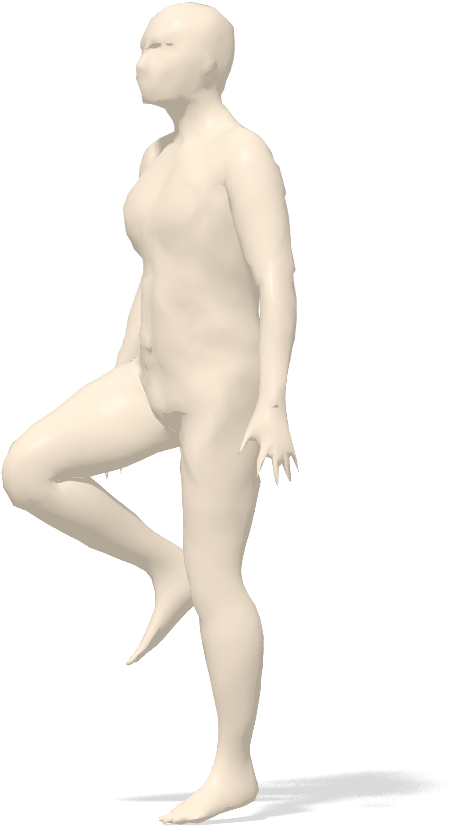}
\put(0,65){\small $\alpha = 1$}
\end{overpic}
\end{tabular}
}
\caption{\textbf{Latent space interpolation.} Interpolation between two poses (left- and right-most shapes) obtained as convex combinations of the respective representations the the latent space. Bottom row: latent space arithmetic.
} 
\label{fig:interpolation}
\end{figure} 
%

\subsection{Completion variability}
\label{subsec:completion_variability}
\begin{figure}
\centering
\resizebox{1.05\columnwidth}{!}{
\begin{tabular}{c@{\hskip 5mm}|@{\hskip 5mm}cccc}
Input & \multicolumn{4}{c}{Completions} \hspace{5mm} \\ \\
\includegraphics[trim=0 0 150 0,clip,height=0.20\textwidth]{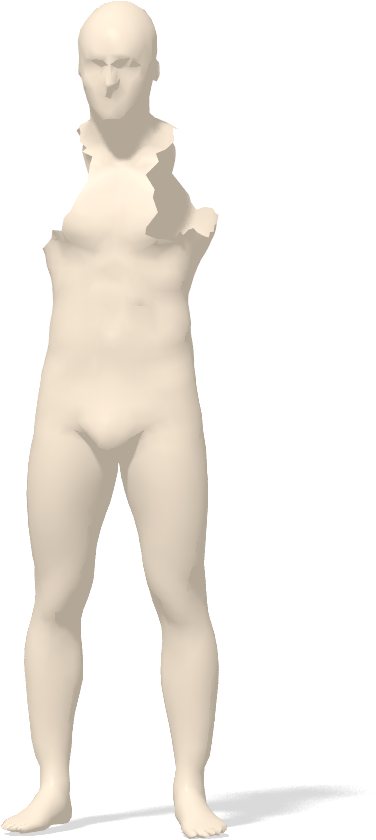} &
\includegraphics[height=0.20\textwidth]{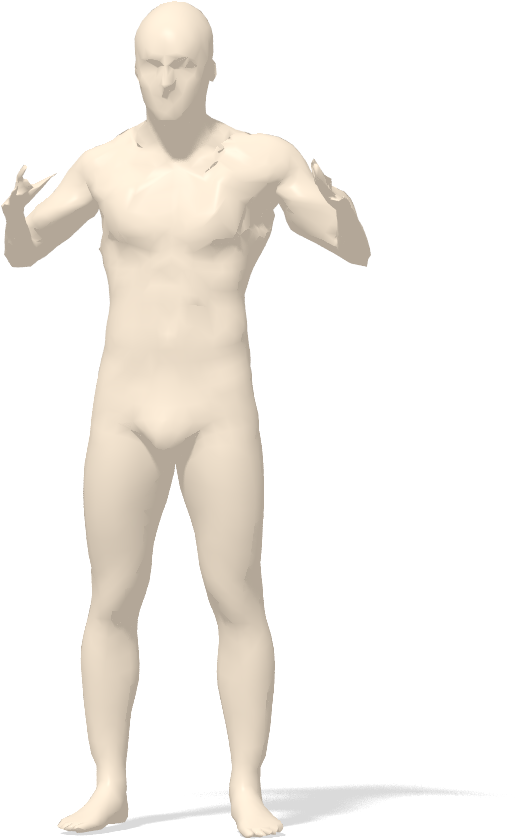} &
\includegraphics[height=0.20\textwidth]{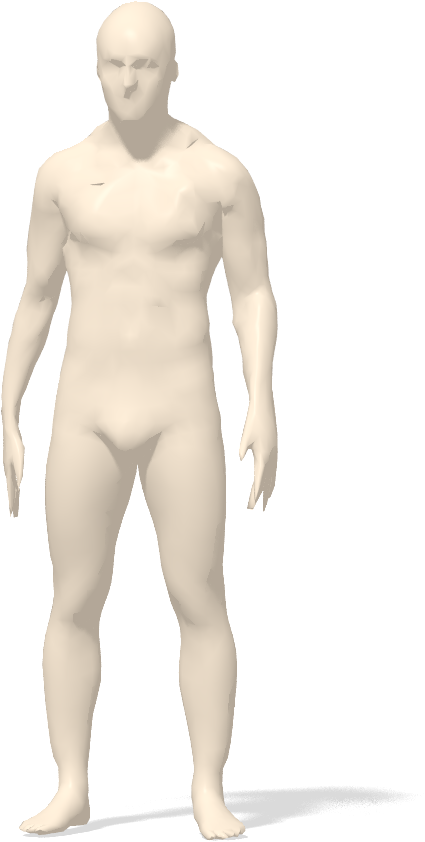} &
\includegraphics[height=0.20\textwidth]{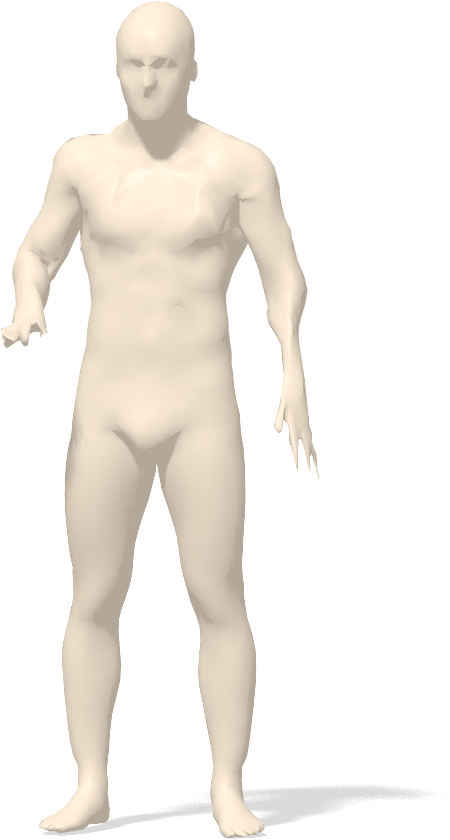} &
\includegraphics[height=0.20\textwidth]{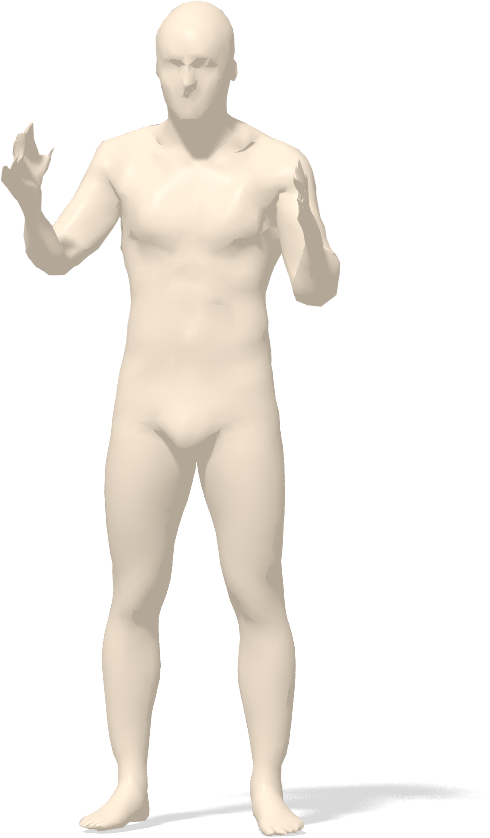} \\
\includegraphics[trim=0 0 150 0,clip,height=0.20\textwidth]{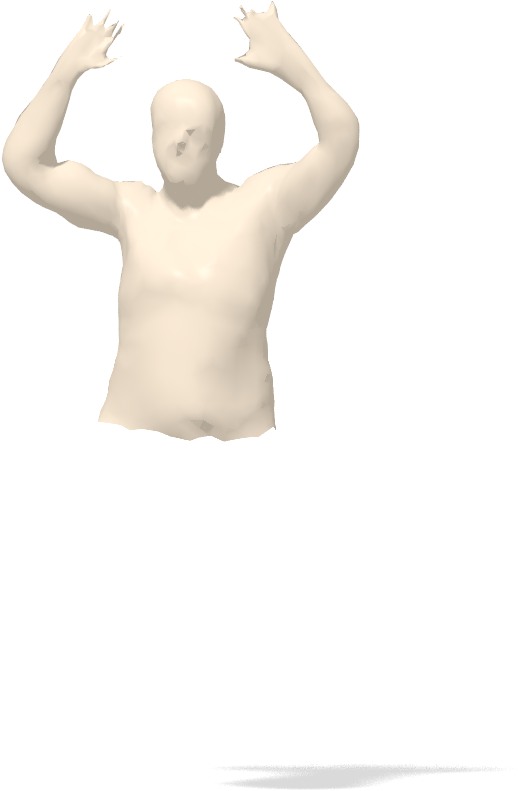} &
\includegraphics[height=0.20\textwidth]{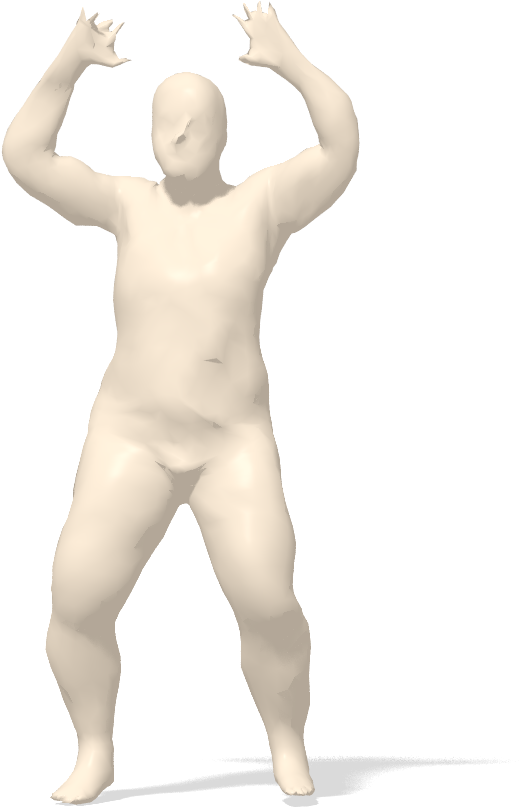} &
\includegraphics[height=0.20\textwidth]{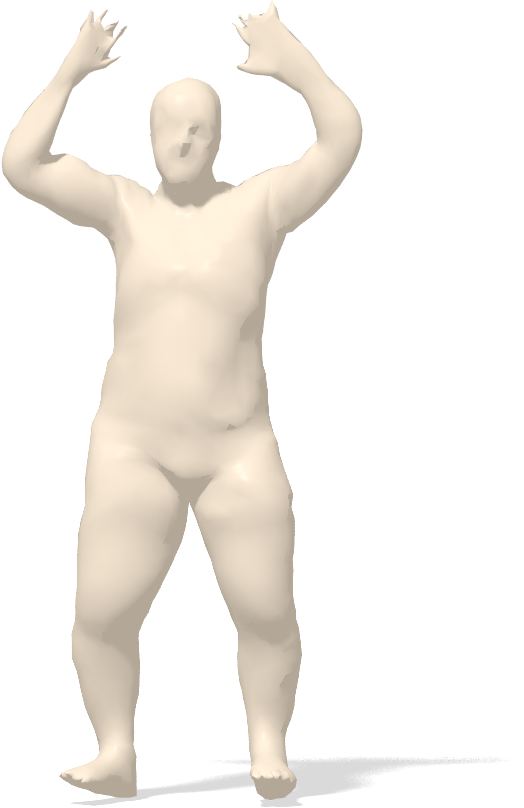} &
\includegraphics[height=0.20\textwidth]{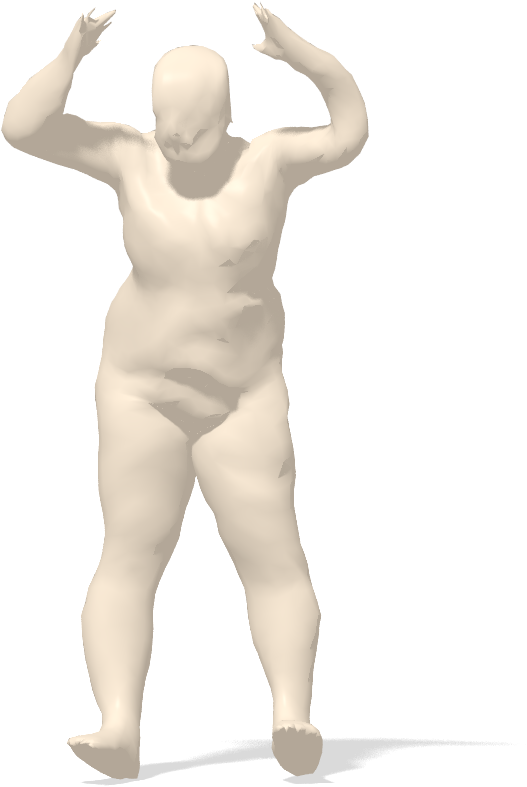} &
\includegraphics[height=0.20\textwidth]{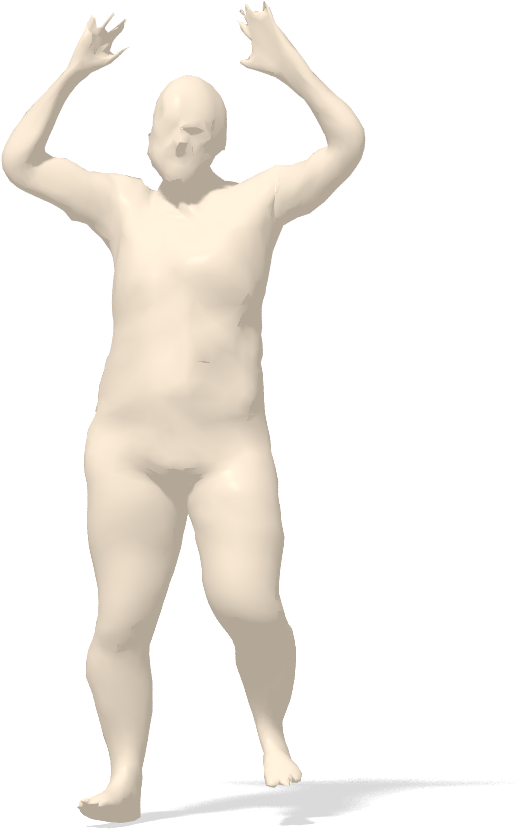} \\
\includegraphics[trim=0 0 150 0,clip,height=0.20\textwidth]{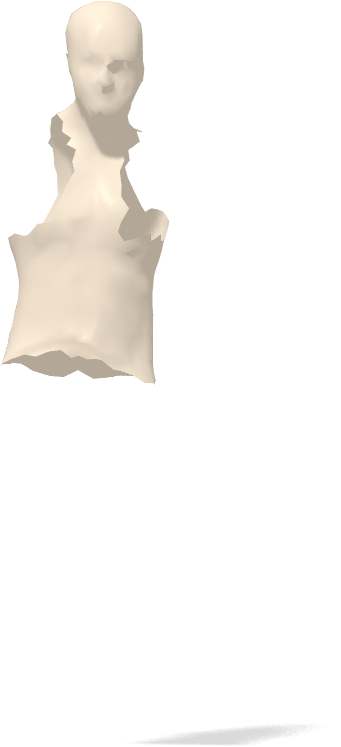} &
\includegraphics[height=0.20\textwidth]{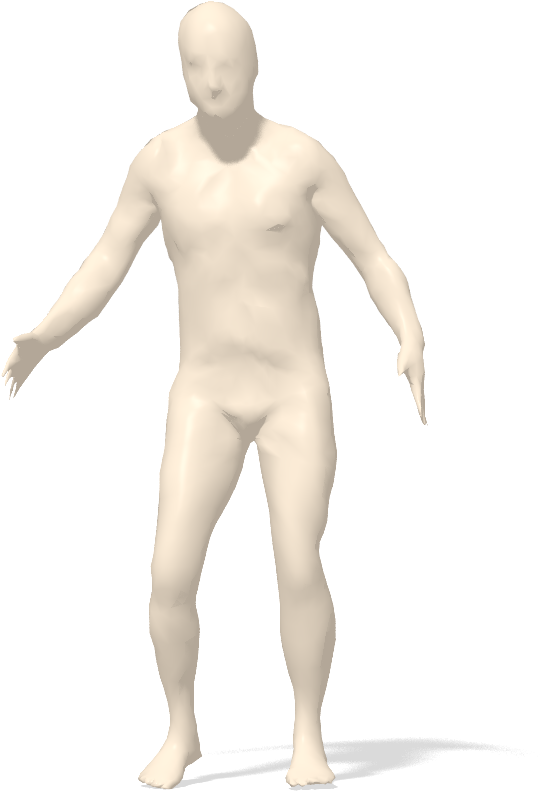} & 
\includegraphics[height=0.20\textwidth]{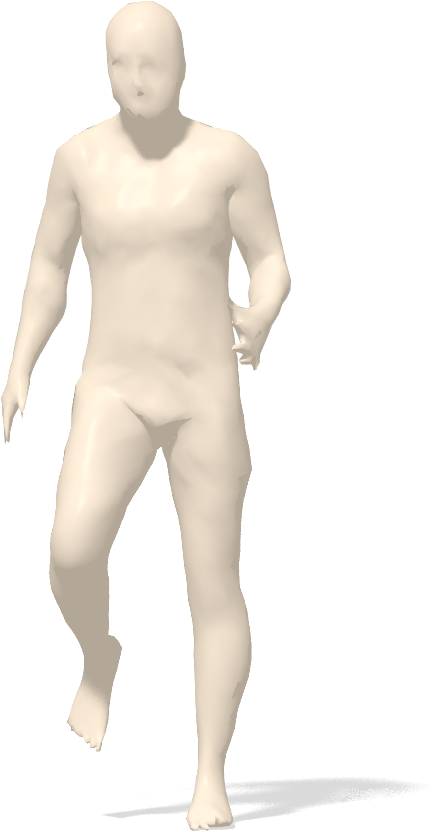} & 
\includegraphics[height=0.20\textwidth]{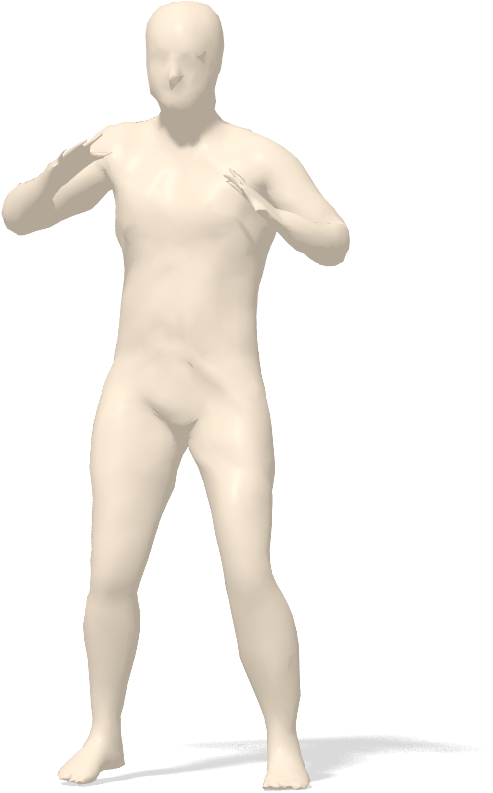} & 
\includegraphics[height=0.20\textwidth]{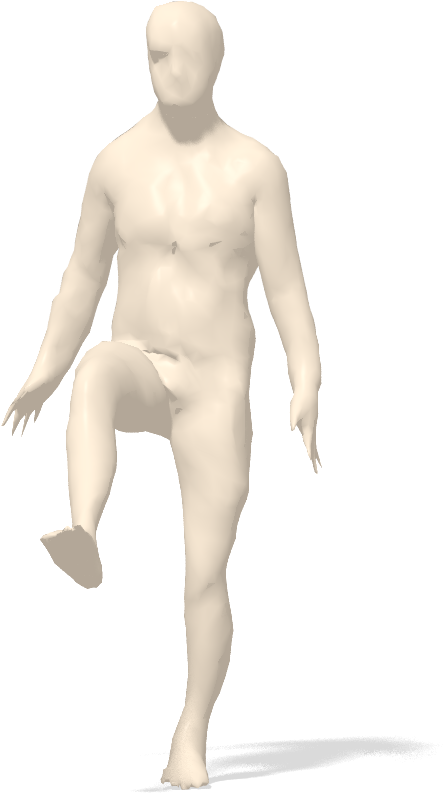}
\end{tabular}
}
\caption{\textbf{Completion variability.} When large contiguous regions (e.g. limbs) are missing, the solution to shape completion is not unique. Shown here are different reconstructions with our method obtained using random initializations.
} 
\label{fig:completion_variability}
\end{figure}

As explained in Section \ref{sec:method}, given a partial input with more than one solution consistent with the data, we may explore this space of completions by sampling the initialization of problem~\ref{eq:optimization} at random from the Gaussian prior. For evaluation we consider several test subjects with removed limbs. Figure \ref{fig:completion_variability} shows unique plausible completions of the same partial input achieved by random initializations.

\definecolor{mygray}{rgb}{0.6,0.6,0.6}

%

\subsection{Synthetic range scans completion}
\label{subsec:range_scans_completion}
\begin{figure}
\centering
\includegraphics[width=0.5\columnwidth]{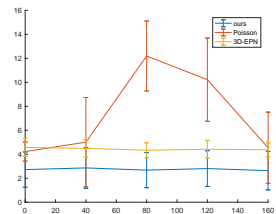}
\caption{\textbf{Reconstruction error as a function of view angle.} Our method produces a consistently accurate reconstruction independent of the view angle. } 
\label{fig:err_matrix}
\end{figure}

\begin{figure*}
\centering
\addtolength{\tabcolsep}{-4pt}
\begin{tabular}{c@{\hskip 7mm}c@{\hskip 7mm}c@{\hskip 7mm}c@{\hskip 7mm}c@{\hskip 7mm}c}
Input & Ground truth & Poisson & NN & 3D-EPN & Ours \\ \\
\includegraphics[trim=0 0 75 0,clip,height=0.18\textwidth]{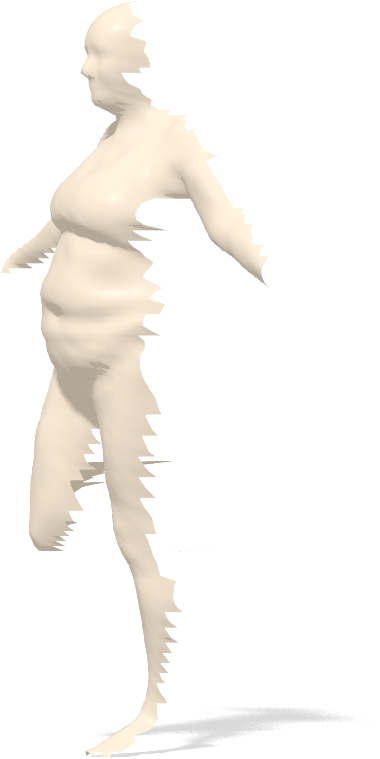} &
\includegraphics[trim=0 0 75 0,clip,height=0.18\textwidth]{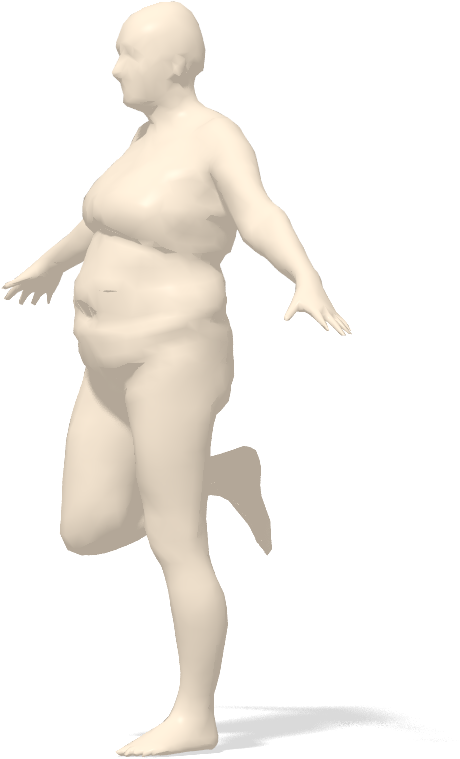} &
\includegraphics[trim=0 0 75 0,clip,height=0.18\textwidth]{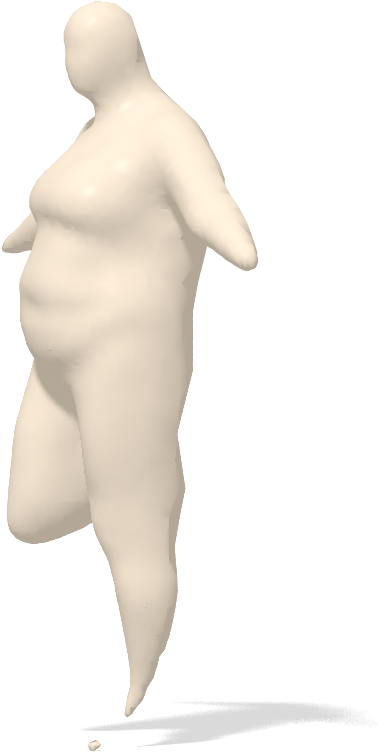} &
\includegraphics[trim=0 0 75 0,clip,height=0.18\textwidth]{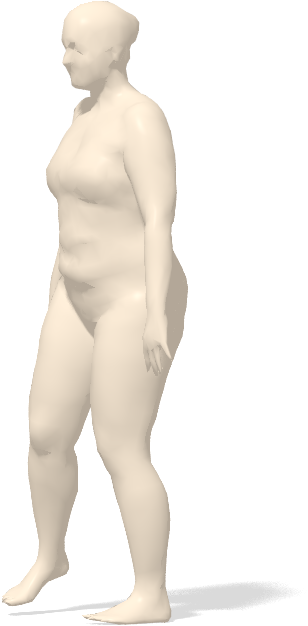} &
\includegraphics[trim=0 0 75 0,clip,height=0.18\textwidth]{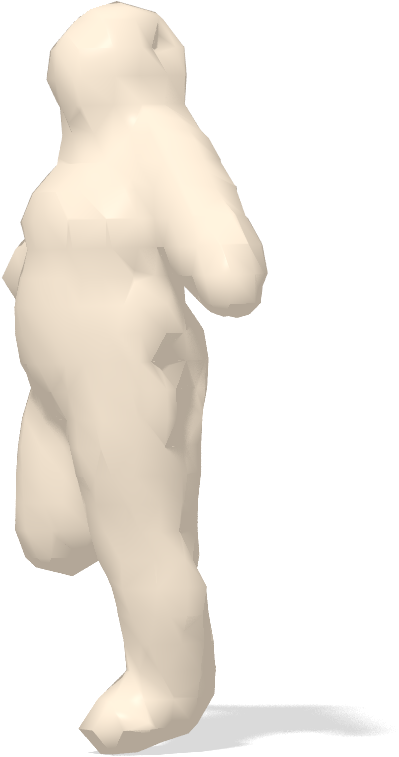} &
\includegraphics[trim=0 0 75 0,clip,height=0.18\textwidth]{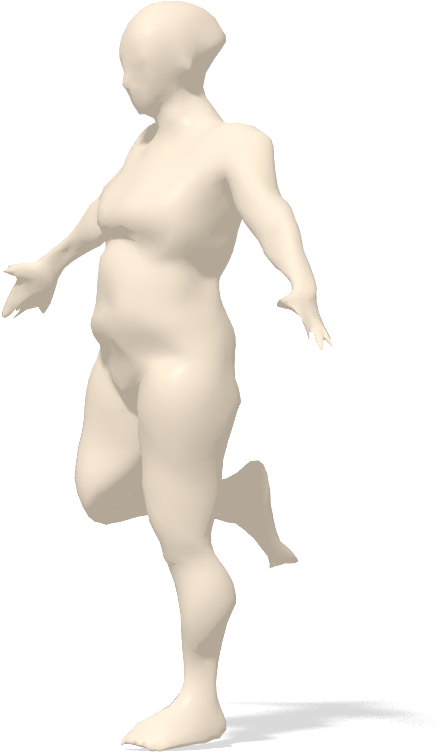} \\ 
\includegraphics[trim=0 0 75 0,clip,height=0.18\textwidth]{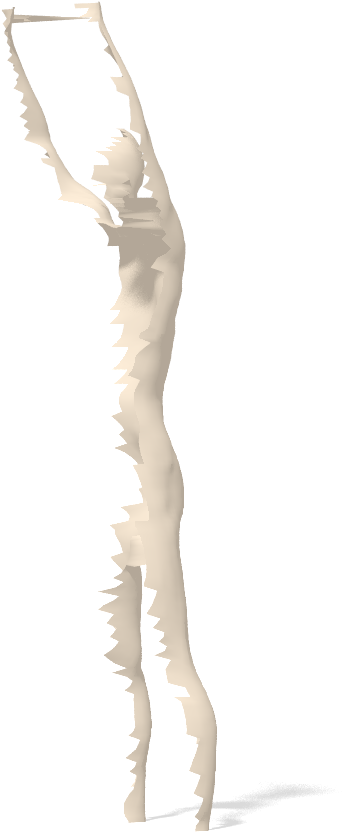} &
\includegraphics[trim=0 0 75 0,clip,height=0.18\textwidth]{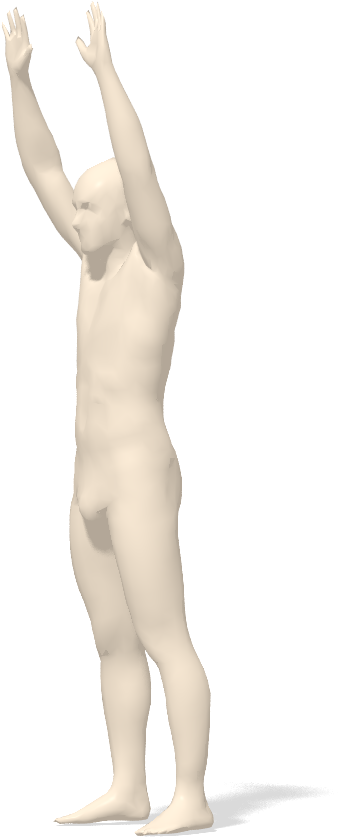} &
\includegraphics[trim=0 0 75 0,clip,height=0.18\textwidth]{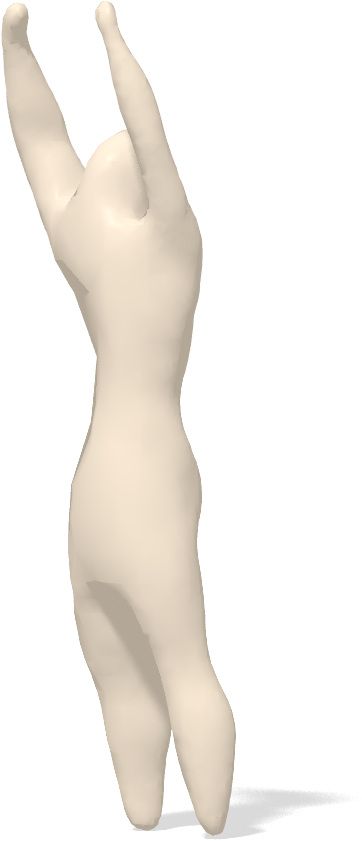} &
\includegraphics[trim=0 0 75 0,clip,height=0.18\textwidth]{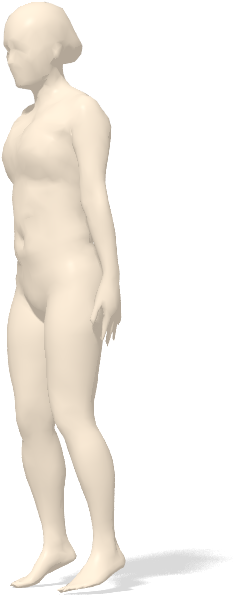} &
\includegraphics[trim=0 0 75 0,clip,height=0.18\textwidth]{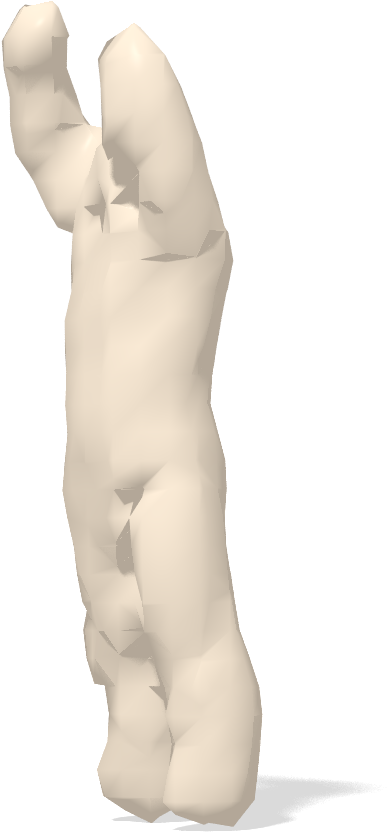} &
\includegraphics[trim=0 0 75 0,clip,height=0.18\textwidth]{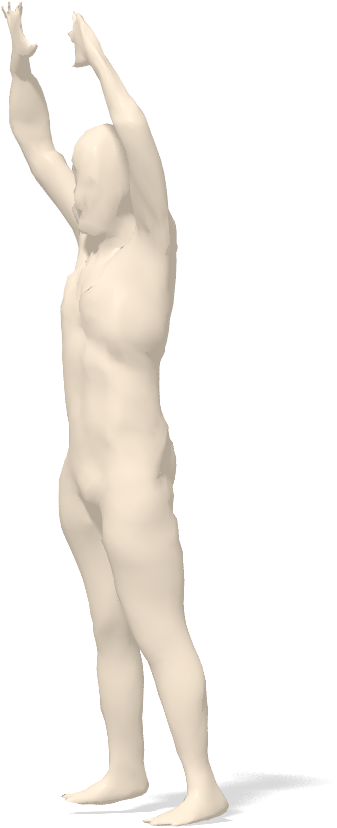}
\end{tabular}
\caption{\textbf{Comparison of different synthetic range scan completion methods.} From left to right: input range scan, ground truth complete shape, Poisson reconstruction, 3D-EPN, and our method. 
} 
\label{fig:proj_completion}
\end{figure*}

The following experiment considers the common practical scenario of range scan completion.
We utilize a test-set of $200$ virtual scans produced from $10$ viewpoints around $2$ human subjects exhibiting $10$ different poses. The full shapes were taken from FAUST~\cite{Bogo:CVPR:2014}, and are completely disjoint from our train set, as they contain novel subjects and poses. Furthermore, the data is suitable for quantitative comparison as sufficient information is given in the partial shape to make the completion problem nearly deterministic. Keeping the ground truth correspondence from each view to the full shape, we report the mean completion error in table~\ref{tab_proj_completion} as \emph{Ours (ground truth)}. More interesting are the results of end-to-end completion using partial correspondence obtained by MoNet~\cite{monet} (reported as \emph{Ours (MoNet)}). For reference we report the performance of other shape completion methods: 3D-EPN \cite{dai2016shape} which has shown state-of-the-art performance for shape completion using volumetric networks, Poisson reconstruction \cite{kazhdan2013screened}, and nearest neighbor (NN). Note, in order to comply with the architecture of 3D-EPN, we also provide viewpoint information, which is unknown for our method. For NN the completion is considered to be the closest shape from the entire training using the ground truth correspondences. Results in table~\ref{tab_proj_completion} show mean Euclidean distance (in cm) and relative volumetric error (in \%) for the missing region. More results are shown in Figures~\ref{fig:err_matrix} and~\ref{fig:proj_completion}.

\paragraph{Robust optimization.}
Our method is able to generalize well to partial shapes in poses unseen during training. However, there is a gap in performance when using correspondences from an oracle versus an off-the-shelf technique (MoNet). To handle noisy correspondences better, we propose a robust enhancement to (\ref{eq:optimization}). Observing that our method may not converge to the ideal completion if the alignment is guided by poor correspondences. However, if the partial shape is somewhat well aligned with the generated shape we can recalculate the correspondence $\bb{\Pi}$ using a simple Euclidean closest-vertex assignment. We find that recalculating when SGD plateaus leads to improved completions (results are reported in Table~\ref{tab_proj_completion} as \emph{Ours (MoNet with refinement)}).
A pleasant side-effect of this refinement step is that our shape completion method can be used to obtain a de-noised (albeit sparser) set  of correspondences (see the supplemental material for analysis).
We also evaluate shape completion when the optimization steps are capped at 300 (reported as \emph{Ours (MoNet with refinement $300$)}) as opposed to running until convergence. Note, for simplicity we did not explore tuning different aspects of the method (learning rate, different reconstruction losses, etc).

\begin{table}[tb]
\centering
\small
\begin{tabular}{ l@{\hskip 0.01\textwidth}c@{\hskip 0.01\textwidth}c@{\hskip 0.01\textwidth}c@{\hskip 0.01\textwidth}c@{\hskip 0.01\textwidth}c@{\hskip 0.01\textwidth}c@{\hskip 0.01\textwidth}c  }
    \hline\hline
    Error & Euclidean & Volumetric err. \\     
          & distance [cm] & mean $\pm$ std [\%] \\ \hline    
    Poisson \cite{kazhdan2013screened}  & $7.3$ 	& $24.8 \pm 23.2$  \\
    NN   (ground truth)   & $5.4$ 	& $34.01 \pm 9.23$  \\   
    3D-EPN \cite{dai2016shape} & $4.43$ & $89.7 \pm 33.8$  \\        
    {\bf Ours} (MoNet)   & $3.40$ 	& $12.51 \pm 11.1$  \\ 
    {\bf Ours} (MoNet with ref. $300$)   & $3.01$ & $10.00  \pm 8.83$ \\ 
    {\bf Ours} (MoNet with refinement)   & $\textbf{2.84}$ 	& $\textbf{9.24}  \pm \textbf{8.62}$ \\ 
    {\bf Ours} (ground truth)   & $\textbf{2.51}$ 	& $\textbf{7.48} \pm \textbf{5.64}$ \\ \hline \hline 
  \end{tabular}   
  \vspace{2mm}
\caption{\small \textbf{Synthetic range scans completion.} Comparison of different methods with respect to errors in vertex position and shape volume. Our method is evaluated using ground truth and MoNet~\cite{monet} correspondences, as well as with and without refinement (details in Section \ref{subsec:range_scans_completion}).
}
\label{tab_proj_completion}
\end{table}

\subsection{Dynamic Fusion}
\label{subsec:dynamic_fusion}

A common use case of depth scanners is object reconstruction form multiple viewpoints. For static scenes, this problem was explored extensively, e.g., in \cite{newcombe2011kinectfusion,niessner2013real,endres2012evaluation}. Non-rigid deformations pose a much bigger challenge. The works of \cite{newcombe2015dynamicfusion} and \cite{innmann2016volumedeform} have shown very impressive reconstructions, however they are limited to small motions between consecutive frames. This limitation was addressed in \cite{slavcheva2017killingfusion} by introducing a damped Killing motion constraint. These methods are focused on reconstructing only the observed dynamic surfaces and cannot convincingly hallucinate large unseen regions. 
Having developed a completion method for non-rigid shapes, we propose its extension to multiple partial inputs. Registering the partial inputs, or the individual reconstructed shapes, is challenging. Instead, we propose to merge shapes in the latent space: we obtain $\bb{z}$ by averaging the completed shape latent variables for each partial input, and $\mathrm{dec}(\bb{z})$ produces the fused 3D shape. Since the latent representation mixes body shape and pose, the reconstructed pose will generally not adhere to any of the input poses, but rather will be an interpolation thereof. 

\begin{figure}
\centering
\resizebox{0.9\columnwidth}{!}{
\addtolength{\tabcolsep}{-4pt}
\begin{tabular}{cccc}
\includegraphics[height=0.25\textwidth]{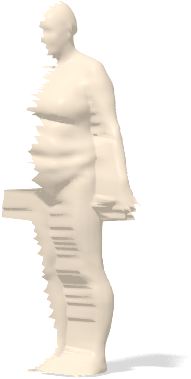} &
\includegraphics[height=0.25\textwidth]{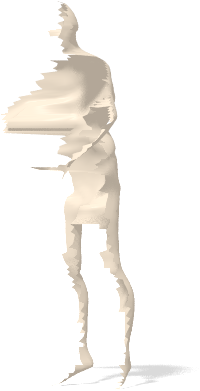} &
\includegraphics[height=0.25\textwidth]{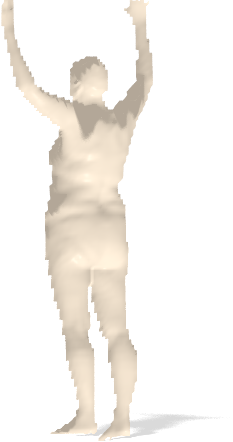} &
\includegraphics[height=0.25\textwidth]{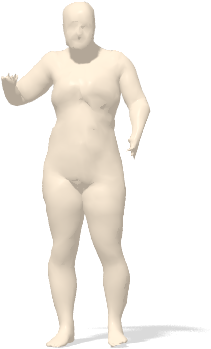} \\
\includegraphics[height=0.25\textwidth]{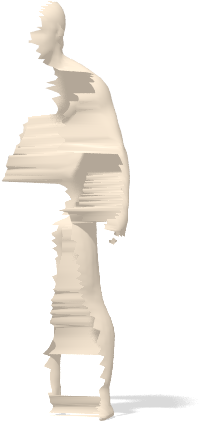} &
\includegraphics[height=0.25\textwidth]{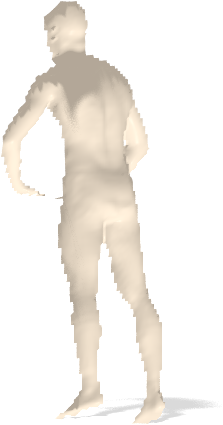} &
\includegraphics[height=0.25\textwidth]{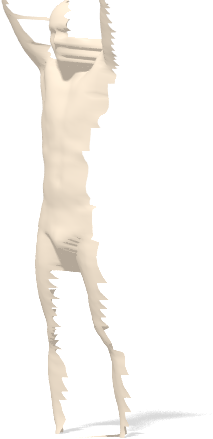} &
\includegraphics[height=0.25\textwidth]{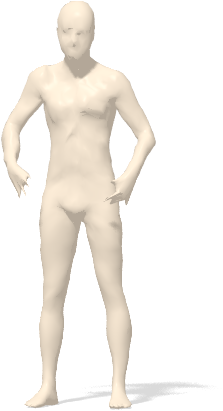}
\end{tabular}
}
\vspace{3mm}
\caption{\textbf{Dynamic fusion.} Three partial views (columns 1-3) and the reconstructed complete shape (rightmost column).} 
\label{fig:dynamic_fusion}
\end{figure}

For a quantitative analysis, we perform fusion on three partial views from a static shape. We use the same FAUST shapes used for testing in Section \ref{subsec:range_scans_completion}. Table~\ref{tab:static_fusion} shows mean reconstruction errors for all $20$ test shapes when fusing three different partial views. The results show how reconstruction accuracy changes according to the viewpoint, and consistently improves with latent space fusion. A qualitative evaluation of the fusion problem is shown for the dynamic setting in Figure~\ref{fig:dynamic_fusion}. Each row shows three partial views of the same human subject from a different viewpoint \emph{and} a different pose. The latent space fusion of the completed shapes is shown in column 4. 

\begin{table}[tb]
\centering
\small
\begin{tabular}{ c c c c }
    \hline\hline
     View 1 & View 2 & View 3 & Fused  \\ \hline  
     $2.78$ & $2.94$ & $2.93$ & $\textbf{2.59}$ \\
     $3.03$ & $3.39$ & $2.73$ & $\textbf{2.61}$ \\ 
     \hline \hline 
\end{tabular}   
  \vspace{2mm}

\caption{\small \textbf{Fusion in the latent space.}
Reported is the mean Euclidean error in cm for three partial views ($0^o$, $120^o$ and $240^o$ for the first row and $80^o$, $200^o$ and $320^o$ for the second row). 
}

\label{tab:static_fusion}
\end{table}

\subsection{Real range scan completion}
The MHAD dataset \cite{ofli2013berkeley} provides Kinect scans from $2$ viewpoints of subjects performing a variety of actions. We apply our completion method to the extracted point cloud (correspondences were initialized through coarse alignment to a training shape, see the supplemental for details). Figure \ref{fig:err_real_data} depicts examples of scan completion on the Kinect data as well as on real scans from the DFAUST dataset.

\begin{figure}
\centering
\resizebox{0.95\columnwidth}{!}{
\begin{tabular}{ccc}
\includegraphics[height=0.3\textwidth]{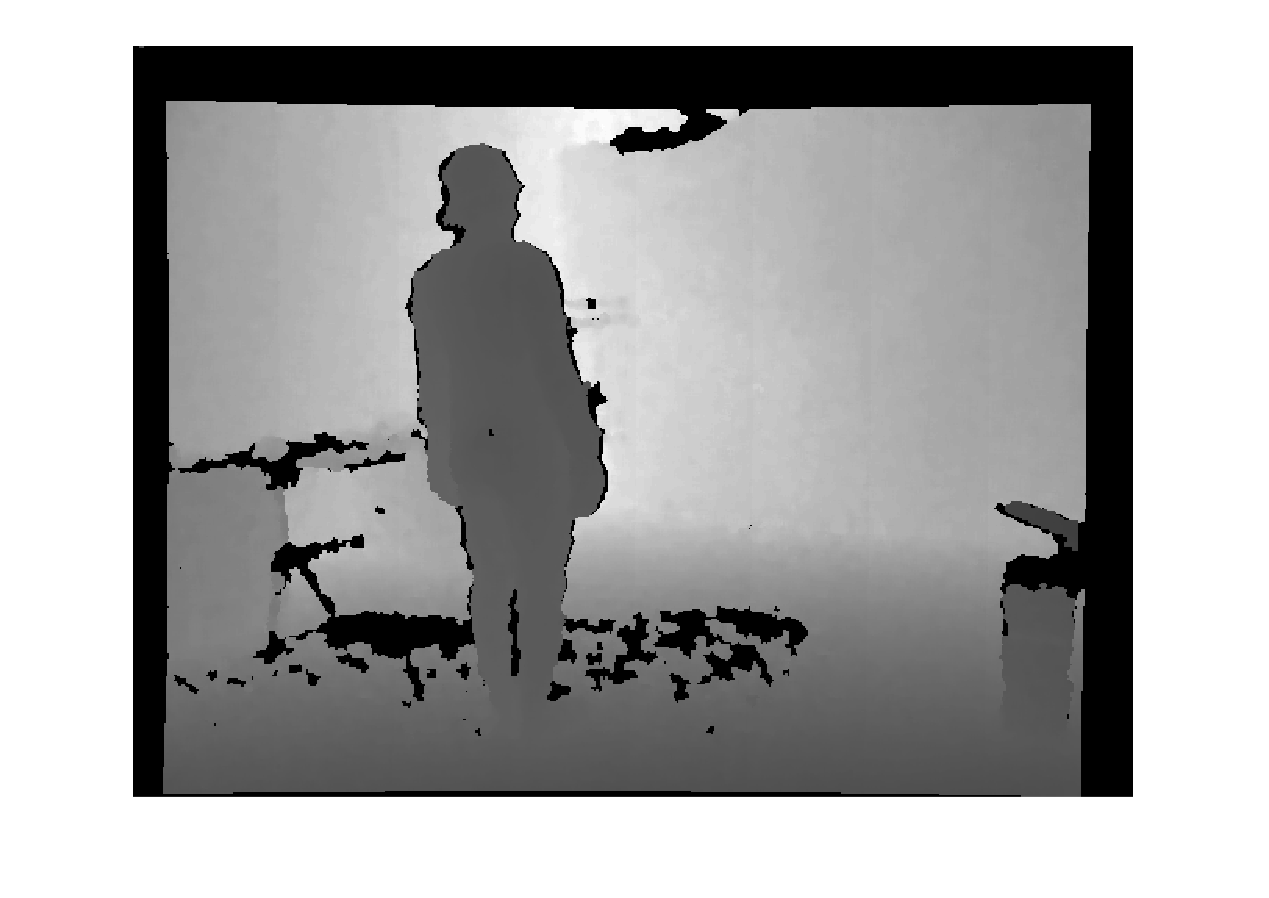} &
\includegraphics[height=0.3\textwidth]{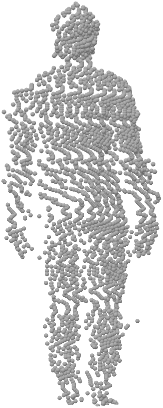} &
\includegraphics[trim=0 0 40 0,clip,height=0.3\textwidth]{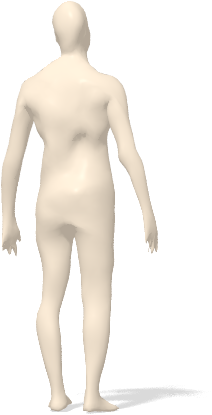} \\
\includegraphics[height=0.3\textwidth]{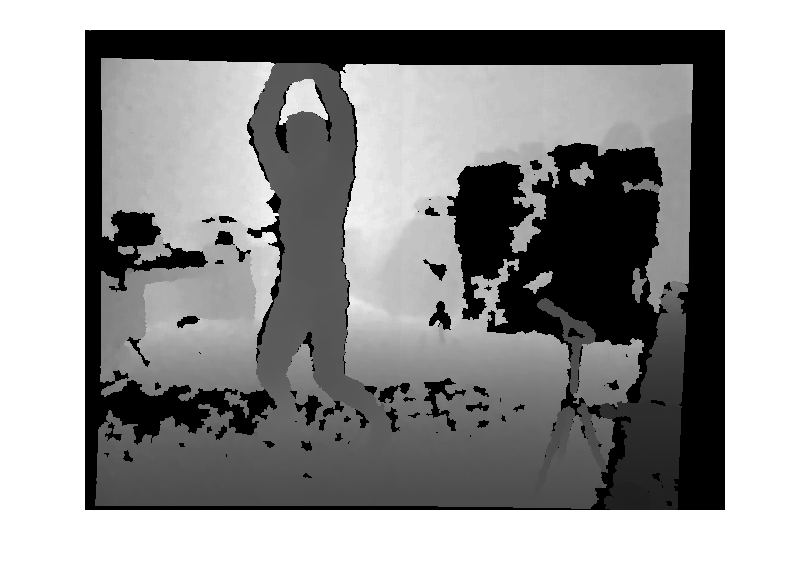} &
\includegraphics[height=0.3\textwidth]{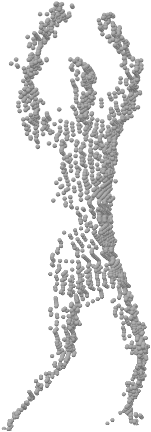} &
\includegraphics[trim=0 0 40 0,clip,height=0.3\textwidth]{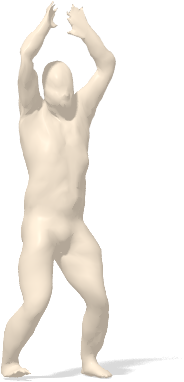}
\end{tabular}
}
\newline
\resizebox{0.95\columnwidth}{!}{
\begin{tabular}{cc@{\hskip 15mm}cc}
\includegraphics[height=0.3\textwidth]{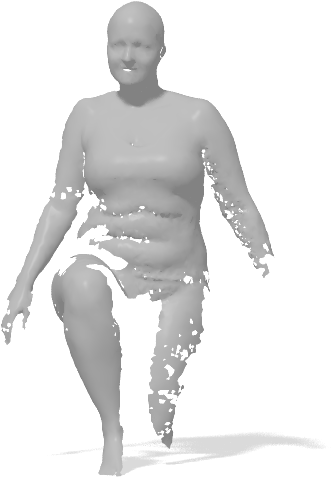} &
\includegraphics[trim=0 0 40 0,clip,height=0.3\textwidth]{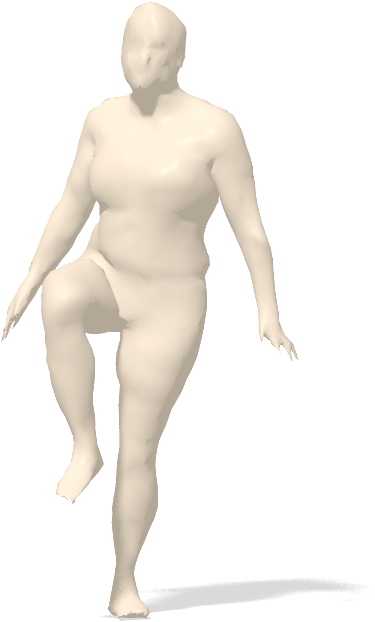} &
\includegraphics[height=0.3\textwidth] {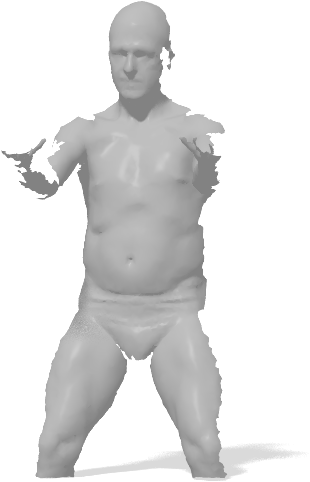} &
\includegraphics[trim=0 0 40 0,clip,height=0.3\textwidth]{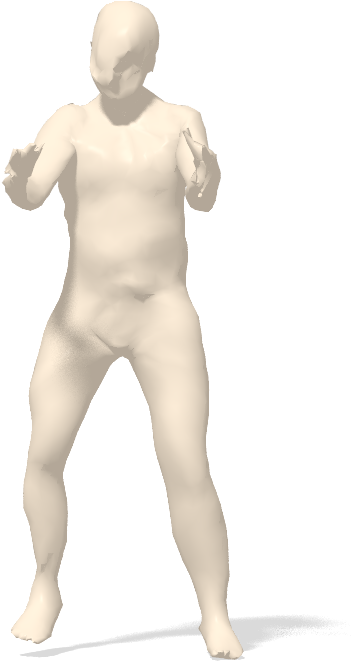}
\end{tabular}
}
\caption{\textbf{Completion of real range scans.} First two rows: completion of Kinect scans (left: depth image; middle: extracted point cloud; right: completed shape). Last row: completion of scans from the DFAUST dataset.
\vspace{-2mm}
} 
\label{fig:err_real_data}
\end{figure} 

\subsection{Face completion}
\label{sec:facecompletion}
\begin{figure}
\centering
\resizebox{0.95\columnwidth}{!}{
\begin{tabular}{cccc}
\includegraphics[width=0.11\textwidth]{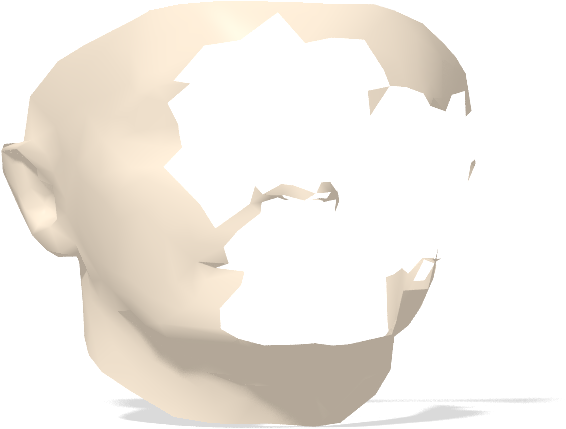} &
\includegraphics[width=0.11\textwidth]{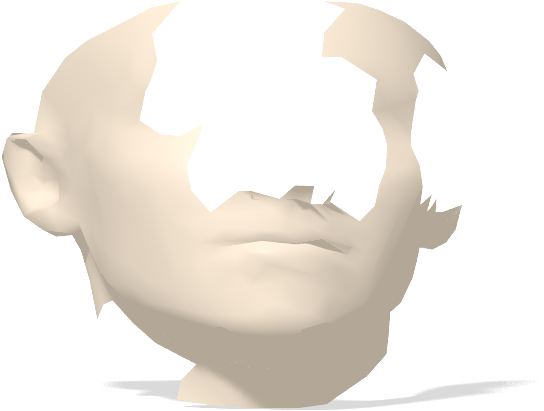} &
\includegraphics[width=0.10\textwidth]{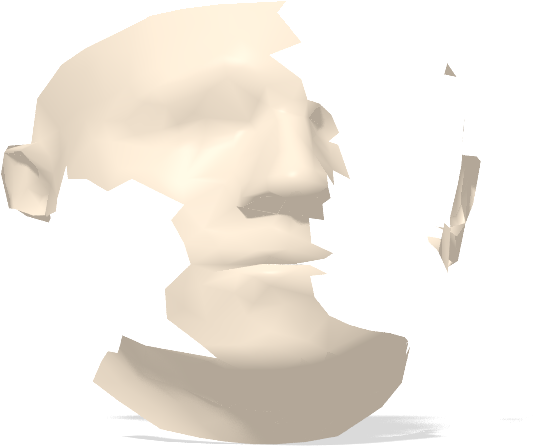} &
\includegraphics[width=0.11\textwidth]{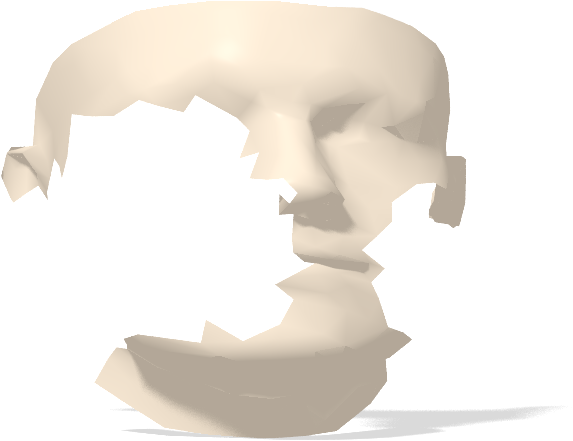} \\

\includegraphics[width=0.11\textwidth]{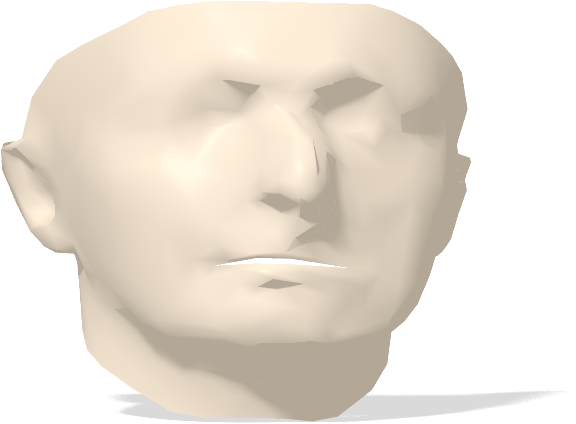} &
\includegraphics[width=0.11\textwidth]{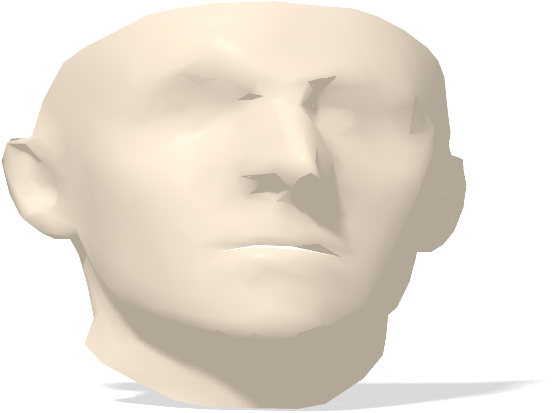} &
\includegraphics[width=0.11\textwidth]{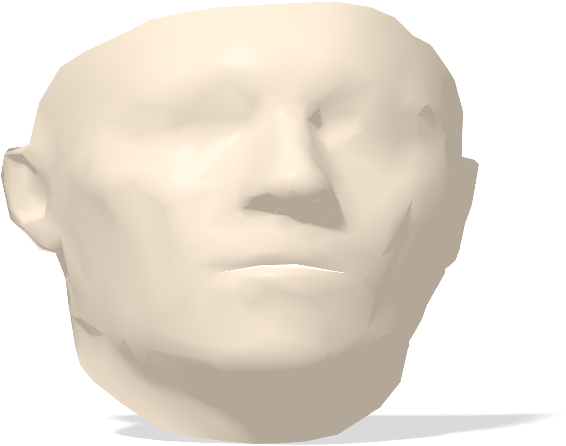} &
\includegraphics[width=0.11\textwidth]{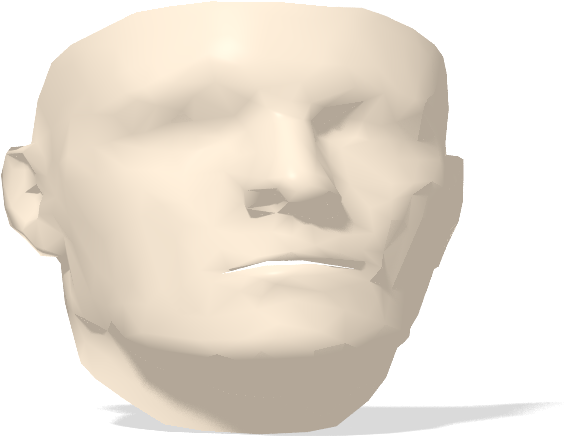} \\ \\

\includegraphics[width=0.11\textwidth]{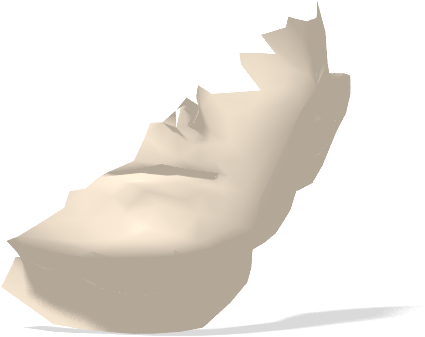} &
\includegraphics[width=0.09\textwidth]{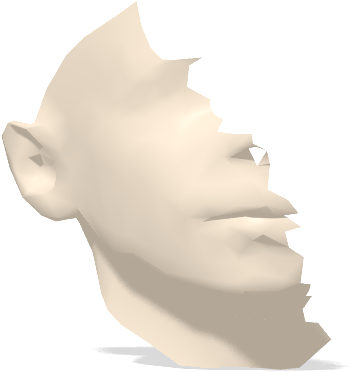} &
\includegraphics[width=0.11\textwidth]{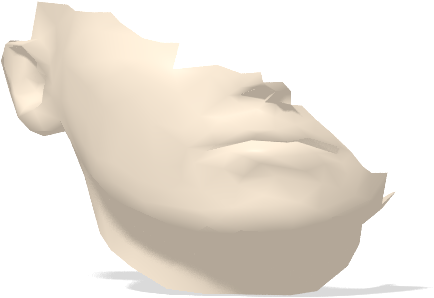} &
\includegraphics[width=0.11\textwidth]{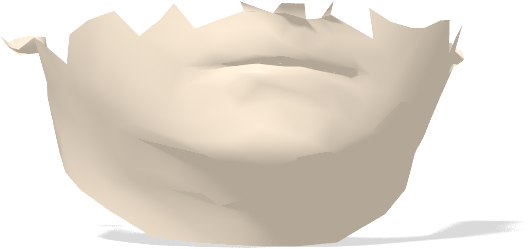} \\

\includegraphics[width=0.11\textwidth]{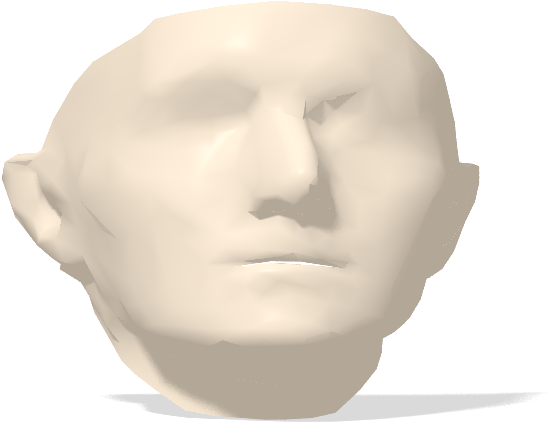} &
\includegraphics[width=0.11\textwidth]{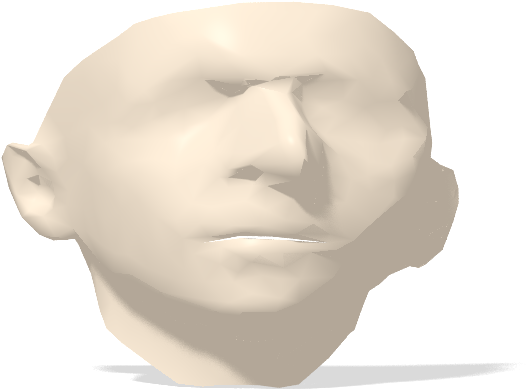} &
\includegraphics[width=0.11\textwidth]{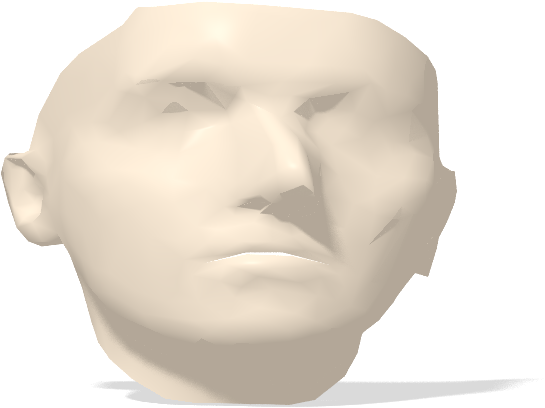} &
\includegraphics[width=0.11\textwidth]{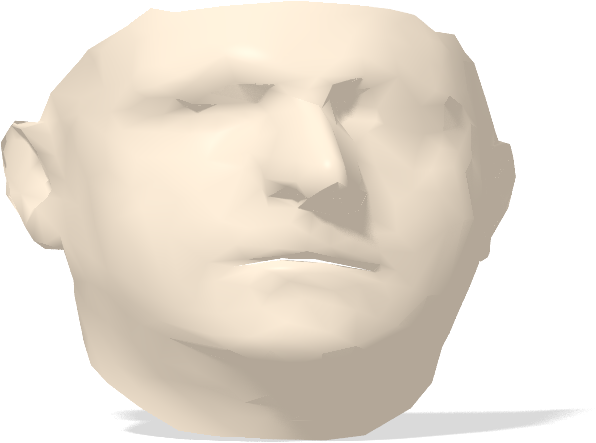} \\ \\

\includegraphics[width=0.11\textwidth]{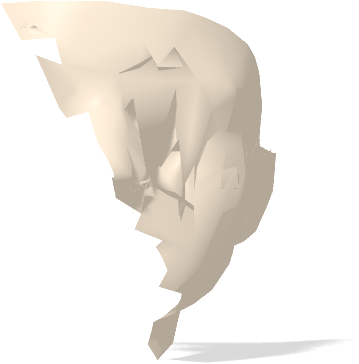} &
\includegraphics[width=0.11\textwidth]{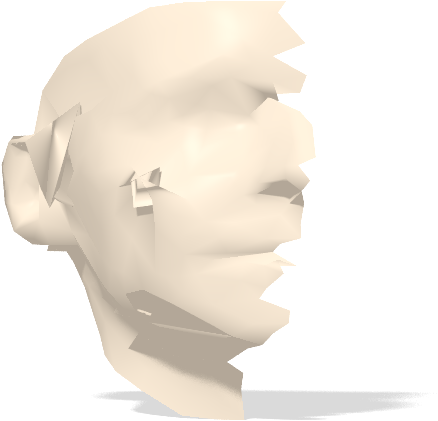} &
\includegraphics[width=0.11\textwidth]{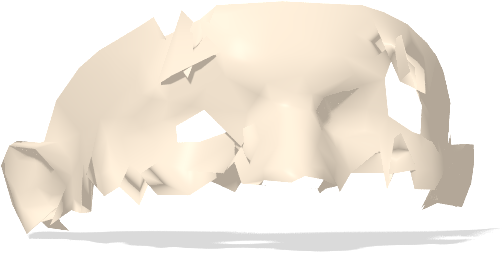} &
\includegraphics[width=0.11\textwidth]{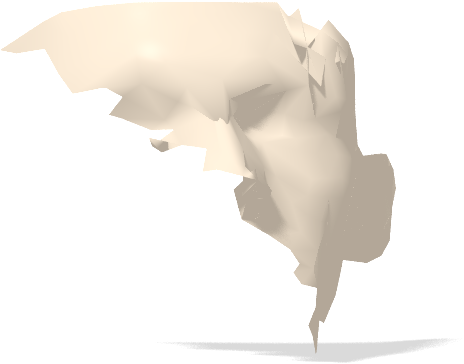} \\ 

\includegraphics[width=0.11\textwidth]{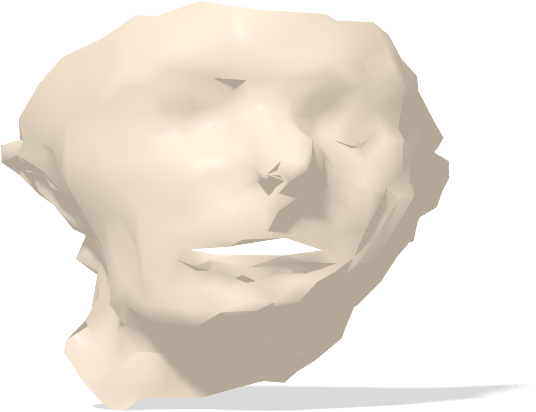} &
\includegraphics[width=0.11\textwidth]{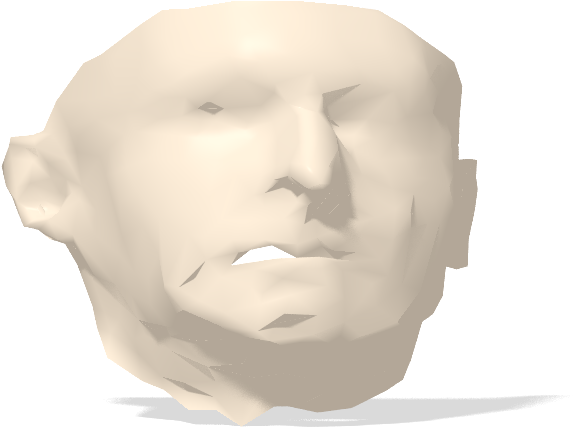} &
\includegraphics[width=0.11\textwidth]{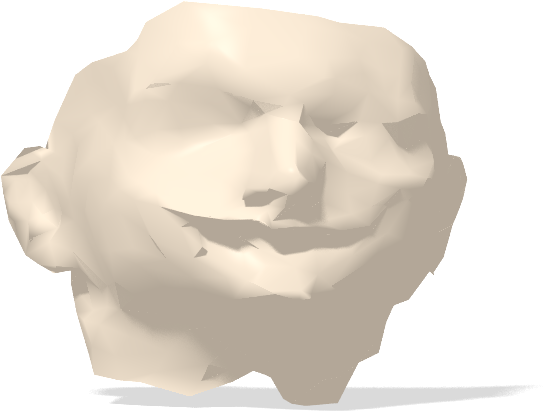} &
\includegraphics[width=0.11\textwidth]{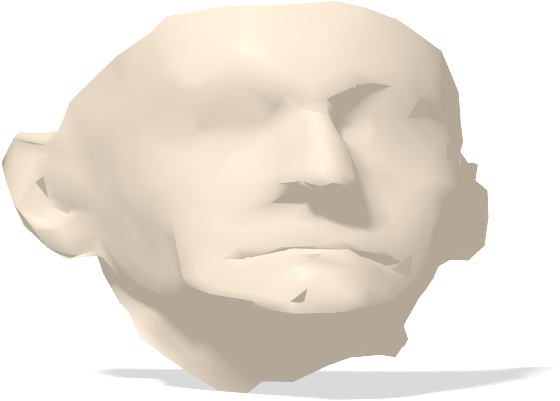} \\ \\

\includegraphics[width=0.11\textwidth]{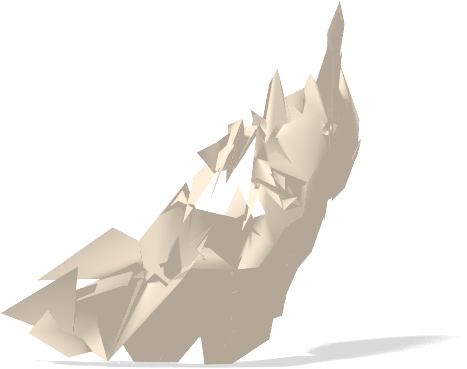} &
\includegraphics[width=0.06\textwidth]{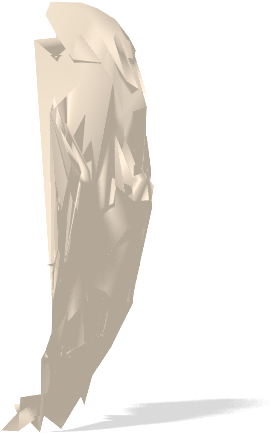} &
\includegraphics[width=0.11\textwidth]{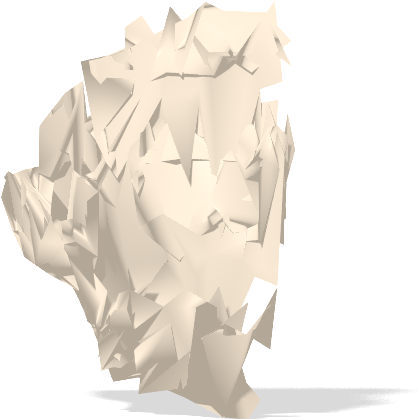} &
\includegraphics[width=0.11\textwidth]{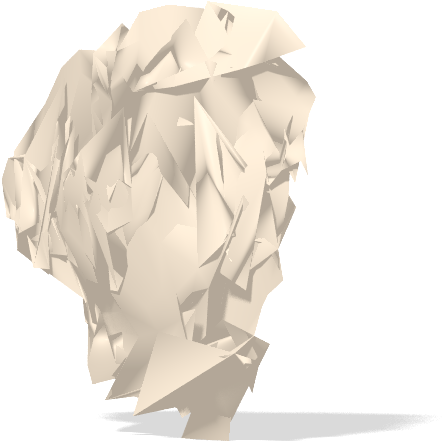} \\ 

\includegraphics[width=0.11\textwidth]{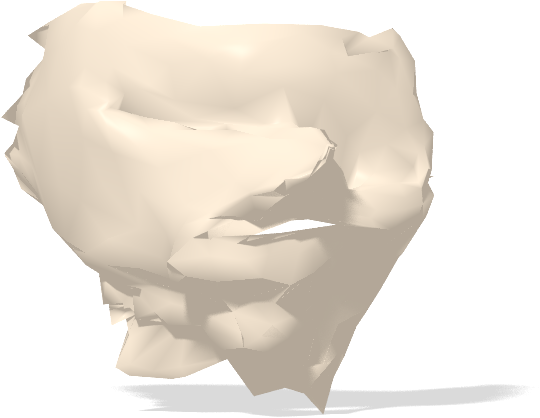} &
\includegraphics[width=0.11\textwidth]{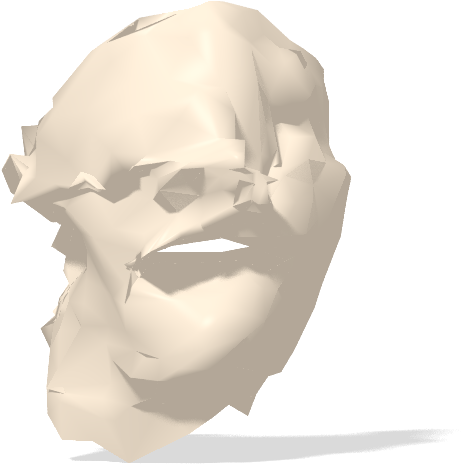} &
\includegraphics[width=0.11\textwidth]{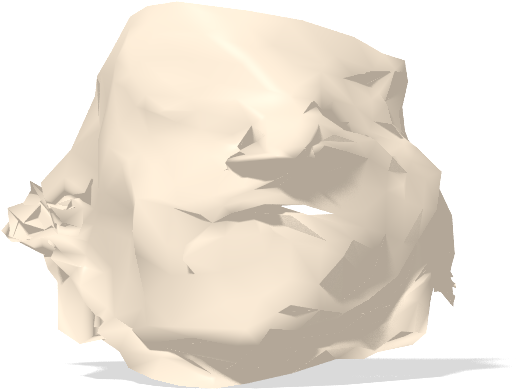} &
\includegraphics[width=0.11\textwidth]{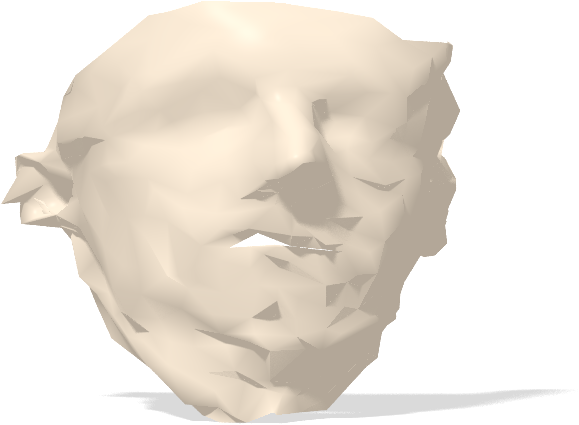}
\end{tabular}
}
\caption{{\bf Completion of faces.} Inputs and outputs are shown in the odd and even rows, respectively.
Rows 1-2 show completion for missing patches, rows 3-4 show completion for hyperplane cuts, rows 5-6 show completion for hyperplane cuts and 5\% correspondence error, and rows 7-8 show 30\% correspondence error. Results indicate completion is plausible even under large missing regions and robust to reasonable correspondence error.}
\label{fig:faces}
\end{figure}

A strength of our fully data-driven approach is that by avoiding explicit shape modeling it generalizes easily to different classes of shapes. This is illustrated by an evaluation on deformable faces. $2000$ training face meshes, each with $525$ vertices, are generated from the model provided by \cite{gerig2017morphable}. These face models exhibit less variability in pose relative to the human meshes, so we use a much smaller VAE network (only two convolutional layers and a latent dimensionality of $32$). Figure~\ref{fig:faces} shows completion for different styles of simulated partiality as well as simulated correspondence noise (see the supplemental for more details).

%% file: conclusions.tex
\section{Conclusions and future work}
This paper introduces a novel graph-convolutional method for shape completion. Its important properties include a model robust to non-rigid deformations, small sample complexity when training, and the ability to reconstruct any style of missing data. Evaluations indicate this is a promising first step towards shape completion from real-world scans, and the analysis reveals directions for future work.
Firstly, exploring a representation that disentangles shape and pose would allow for more control in the completion and likely improve dynamic fusion results.
Secondly, for initialization we require correspondences between the partial and canonical shape model. Although we show resilience to poor correspondences, improving this initialization for noisy real-world data would be beneficial.
Finally, the proposed formulation assumes the desired shape topology (i.e. vertex connectivity) is known when decoding shapes. We leave to future work the task of completion with unknown topology.